\documentclass[10pt,journal,compsoc]{IEEEtran}

\ifCLASSOPTIONcompsoc

  \usepackage[nocompress]{cite}
\else

  \usepackage{cite}
\fi

\ifCLASSINFOpdf

\else

\fi

\usepackage{graphicx}
\usepackage{physics}
\usepackage{amssymb}
\usepackage{xcolor}
\usepackage{bm}
\usepackage{placeins}
\usepackage{multirow} 
\usepackage[hidelinks]{hyperref}

\begin{document}

\title{A Semantic and Motion-Aware Spatiotemporal Transformer Network for Action Detection}

\author{Matthew~Korban, ~\IEEEmembership{Senior Member,~IEEE}, Peter~Youngs,
        Scott~T.~Acton, ~\IEEEmembership{Fellow,~IEEE}

\thanks{\emph{This material is based upon work supported by the National Science Foundation under Grant No. 2000487 and partly supported by NSF under Grant No. 2322993.} (Corresponding author: Scott T. Acton.)

\emph{Matthew Korban and Scott T. Acton are with the Department
of Electrical and Computer Engineering, University of Virginia, Charlottesville, VA 22904 (e-mail: \{acw6ze@virgnia.edu, acton@virginia.edu\}).}

 \emph{Peter Youngs is with the Department of Curriculum, Instruction and Special Education at the University of Virginia, Charlottesville, VA 22903 (email: pay2n@virginia.edu).}}}

\markboth{Journal of \LaTeX\ Class Files}%
{Shell \MakeLowercase{\textit{et al.}}: Bare Demo of IEEEtran.cls for Computer Society Journals}

\IEEEtitleabstractindextext{%
\begin{abstract}
This paper presents a novel spatiotemporal transformer network that introduces several original components to detect actions in untrimmed videos. First, the multi-feature selective semantic attention model calculates the correlations between spatial and motion features to model spatiotemporal interactions between different action semantics properly. Second, the motion-aware network encodes the locations of action semantics in video frames utilizing the motion-aware 2D positional encoding algorithm. Such a motion-aware mechanism memorizes the dynamic spatiotemporal variations in action frames that current methods cannot exploit.  
Third, the sequence-based temporal attention model captures the heterogeneous temporal dependencies in action frames. In contrast to standard temporal attention used in natural language processing, primarily aimed at finding similarities between linguistic words, the proposed sequence-based temporal attention is designed to determine both the differences and similarities between video frames that jointly define the meaning of actions. The proposed approach outperforms the state-of-the-art solutions on four spatiotemporal action datasets: AVA 2.2, AVA 2.1, UCF101-24, and EPIC-Kitchens. 
\end{abstract}

}

\maketitle

\IEEEdisplaynontitleabstractindextext

\IEEEpeerreviewmaketitle

\IEEEraisesectionheading{\section{Introduction}}
\label{sec:intro}

\IEEEPARstart{S}{patiotemporal} action detection aims to localize action class instances in untrimmed videos in both spatial and temporal dimensions \cite{oikonomopoulos2010TIP}. However, such a spatiotemporal action detection faces several challenges, including (1) complex spatiotemporal interactions between action semantics, (2) dynamic spatiotemporal variations in action semantics, and (3) heterogeneous temporal dependencies between action frames. We will explain our proposed solutions to solve the challenges above as follows:

The semantics are the meaningful components of actions that can be categorized into two fundamental types: \emph{persons} and \emph{objects}. \cite{tran2012NIPS} is one of the first methods incorporating action semantics into spatiotemporal action detection. However, their method was limited to finding the motion trajectory of localized objects over video frames, which alone might not be enough to model various actions. \cite{weinzaepfel2015ICCV} addressed such a limitation by including persons and objects that better represent human actions. Yet, they did not consider the interactions between semantics, which are crucial parts of actions.
To solve this issue, \cite{girdhar2019CVPR} proposed a video action transformer network that captures the correlations between a central person and the surrounding pixels. However, they did not explicitly model the interactions between all the action semantics in video frames. However, many actions are defined based on the spatiotemporal interactions between action semantics. For example, the action ``kicking a ball'' is characterized by the interaction between a ``person'' and a moving ``ball'', in both spatial and temporal domains. These domains are associated with the spatial and motion properties of action semantics, respectively. More examples will be illustrated in Fig. \ref{Fig:MFE}.  To address this issue, we propose a multi-feature semantic attention model that enables the transformer network to selectively capture the spatiotemporal interactions between action semantics based on the correlations between their motion and spatial features. More details about the multi-feature selective attention model are explained in Section \ref{SEC:MFSSA}.

The positional encoding of the transformer network extends its capability to encode the order information in the data \cite{li2020ATIP}. As an alternative to the sequential connection in RNN and LSTM, the positional encoding allows the transformer network to process sequential data more effectively and efficiently. The positional encoding was initially proposed to represent 1D temporal order information in linguistic words \cite{vaswani2017NIPS}. It is also used to encode spatial order information in computer vision \cite{carion2020ECCV, Dosovitskiy2020ICLR} recently. Two main strategies to incorporate positional information in computer vision have been ordered pixels \cite{carion2020ECCV} and patches of pixels \cite{Dosovitskiy2020ICLR}. 

Nevertheless, the current positional encoding strategies have two issues when dealing with video frames: First, they are still based on standard 1D temporal positional encoding designed for linguistic words, which might not work well for images due to their 2D nature. Furthermore, existing positional encoding algorithms are limited to static spatial order, which cannot accurately represent dynamic action semantics.
Specifically, the positional information of action semantics changes based on their spatiotemporal variations caused by the dynamic movements of action semantics in videos. These spatiotemporal variations will be illustrated later in Fig. \ref{Fig:2DPOS}. Hence, we propose a motion-aware 2D positional encoding algorithm, which is more effective than the standard methods in modeling the positions of action semantics considering their spatiotemporal variations. The motion-aware 2D positional encoding will be discussed in more detail in Section \ref{Sec:MMM}.

Temporal order information is essential in modeling actions since human action is a sequential process over different times \cite{korban2020ECCV}. Human actions are, however, temporally heterogeneous, making modeling the temporal dependencies between action frames challenging. For example, in some actions such as ``running'', the temporal dependencies between similar and adjacent frames are essential. On the other hand, in many actions such as ``jumping'', the critical temporal dependencies are between distinctive and non-adjacent frames, so-called keyframes such as ``start'', ``middle'', and ``end'' of the jump \cite{korban2023PR2}. 

Traditional sequential methods such as Recurrent Neural Network (RNN) encounter difficulty computing heterogeneous temporal dependencies in actions because of limited temporal receptive fields and bias toward adjacent action frames. A better way to model temporal dependencies in actions is using a transformer network, which can more effectively process non-adjacent frames due to a larger temporal receptive field. However, the standard transformer network was initially designed for natural language processing in which the highest temporal dependencies are between the exact linguistic words, followed by words from similar categories \cite{vaswani2017NIPS}. Hence, while the transformer network can include the temporal dependencies between non-adjacent frames, it is still heavily biased toward similar and often adjacent action frames. Consequently, the transformer network struggles to capture the heterogeneous temporal dependencies between distinctive and non-adjacent frames. Therefore, to resolve this issue, we propose a sequence-based temporal attention model to capture heterogeneous temporal dependencies within the transformer network effectively. More details about the above are discussed in Section \ref{SEC:PTA}.

In summary, the main \textbf{contributions} of this paper are as follows:

\begin{itemize}
\item A novel spatiotemporal transformer network is proposed that includes several original components to detect actions in untrimmed videos by properly modeling the action semantics, their interactions, and movements. The new design solves the fundamental issues of the standard transformer network in action modeling and understanding. To do this, the transformer network and the attention mechanism are renovated in both spatial and temporal domains.
\item We are the first to develop a multi-feature selective semantic attention model to capture the important spatiotemporal interactions between action semantics based on their spatial and motion properties correlations. The multi-feature attention handles the limitation of the standard self-attention, which is restricted to a single feature space. The proposed transformer also uses a selective attention model, which, by selecting informative inputs, is more effective and efficient than the standard self-attention in modeling action semantics.
\item A novel motion-aware 2D positional encoding is introduced that uses a 2D motion memory module to model dynamic spatiotemporal semantic variations in action frames. This overcomes the drawback of the standard 1D positional encoding, which is limited to static spatial positions, making it ineffective in dealing with 2D images and movements in videos.  \item A sequence-based temporal attention model is suggested that, along with the sequence-based temporal positional encoding, can effectively capture heterogeneous temporal dependencies in action frames that might exist in distinctive and often non-adjacent frames.  This model eliminates the limitations of traditional temporal attention, which is biased toward temporal dependencies between similar and commonly adjacent frames in action videos. 
\item The proposed method outperforms state-of-the-art methods on four public spatiotemporal action benchmarks, AVA (V2.2 and V2.1), UCF101-24, and EPIC-Kitchens.
\end{itemize}

\section{Related Work}
There are three relevant topics covered in this section, including methods based on (1) action semantics, (2) multiple features for action modeling, (3) spatiotemporal action detection.

\subsection{Action Semantics}
To represent action semantics, earlier approaches used handcrafted features, such as shape descriptors extracted from human silhouettes \cite{guo2013TIP} and local binary patterns captured from human parts \cite{khan2014TIP}.
With advances in deep learning, action semantics can now be analyzed more effectively. 
Several studies in the literature used deep networks for action semantics that focus either on persons \cite{kalogeiton2017ICCV} or objects \cite{wang2020PAMI}.
In \cite{kalogeiton2017ICCV}, a convolutional feature-based action tubelet detector is presented for modeling persons over different time periods.
\cite{wang2020PAMI} suggested an object-centric feature alignment mechanism to signify critical objects in action videos. However, various actions in the wild rely on complex interactions between persons and objects.  
To accommodate this complexity, \cite{pan2021CVPR} introduced a transformer network to compute the interactions between persons and objects based on their spatial features. As of yet, no effective strategy has been developed to capture the spatiotemporal relationship between action semantics based on spatial and motion features.

\subsection{Multi-feature Networks in Action Modeling}
\cite{simonyan2014NIPS} introduced one of the first deep networks that used both RGB images and optical flow fields to extract spatial and motion features for action modeling and recognition.   
\cite{feichtenhofer2016CVPR} redesigned its architecture, including softmax and pooling layers, in order to improve a multi-feature network.
Despite taking advantage of multiple features in \cite{simonyan2014NIPS, feichtenhofer2016CVPR}, they did not suggest any explicit solution to employ multiple features effectively in action recognition.
Therefore, \cite{wang2017MM} proposed directly enhancing the individual and hybrid representations of multiple features via a spatiotemporal pyramid pooling and a fusion mechanism, respectively.
 \cite{su2020PAMI} improved this multi-feature representation by using region proposals rather than the whole frames suggested in \cite{simonyan2014NIPS, feichtenhofer2016CVPR, wang2017MM}, to emphasize more informative parts of the optical flow and RGB images.
\cite{Hu2022PAMI} further strengthened the joint representation of multiple features by computing the relationship between feature vectors with a self-attention mechanism. 
Still, calculating the relationship between action semantics based on the correlations between their spatial and motion features remains unresolved. 

\subsection{Spatiotemporal Action Detection}
In one of the earliest spatiotemporal action detection approaches, \cite{oikonomopoulos2010TIP} proposed codebooks in which one codeword represented spatial and temporal information about each action class.
A major weakness of \cite{oikonomopoulos2010TIP} is its inability to successfully detect actions in complex backgrounds due to using handcrafted features.
\cite{weinzaepfel2015ICCV} addressed this shortcoming by developing a deep CNN that could more accurately detect spatiotemporal action instances in complex background environments.
\cite{singh2017ICCV} enhanced such as a deep network in \cite{weinzaepfel2015ICCV} by detecting multiple spatial action instances per frame with a more efficient single-shot multi-box detector. 
Before now, each action proposal was based on a fixed model, accumulating the error over time.
In \cite{yang2019CVPR}, this issue was tackled by proposing a progressive learning framework that adapts to new relevant action contexts as they arise.
As an extension of the previous work, \cite{singh2023WCACV} used the track-of-interest alignment technique to cope with large spatial variations of in-the-wild action videos.
A more effective strategy to tackle both large spatial and temporal variations is the transformer network, which has increasingly been used for action detection and recognition.

\section{Methodology}
\subsection{Method Overview}
\label{Sec:Overview}
\begin{figure*}[h!tbp]
	\centering
	\includegraphics[height=0.43\textwidth]{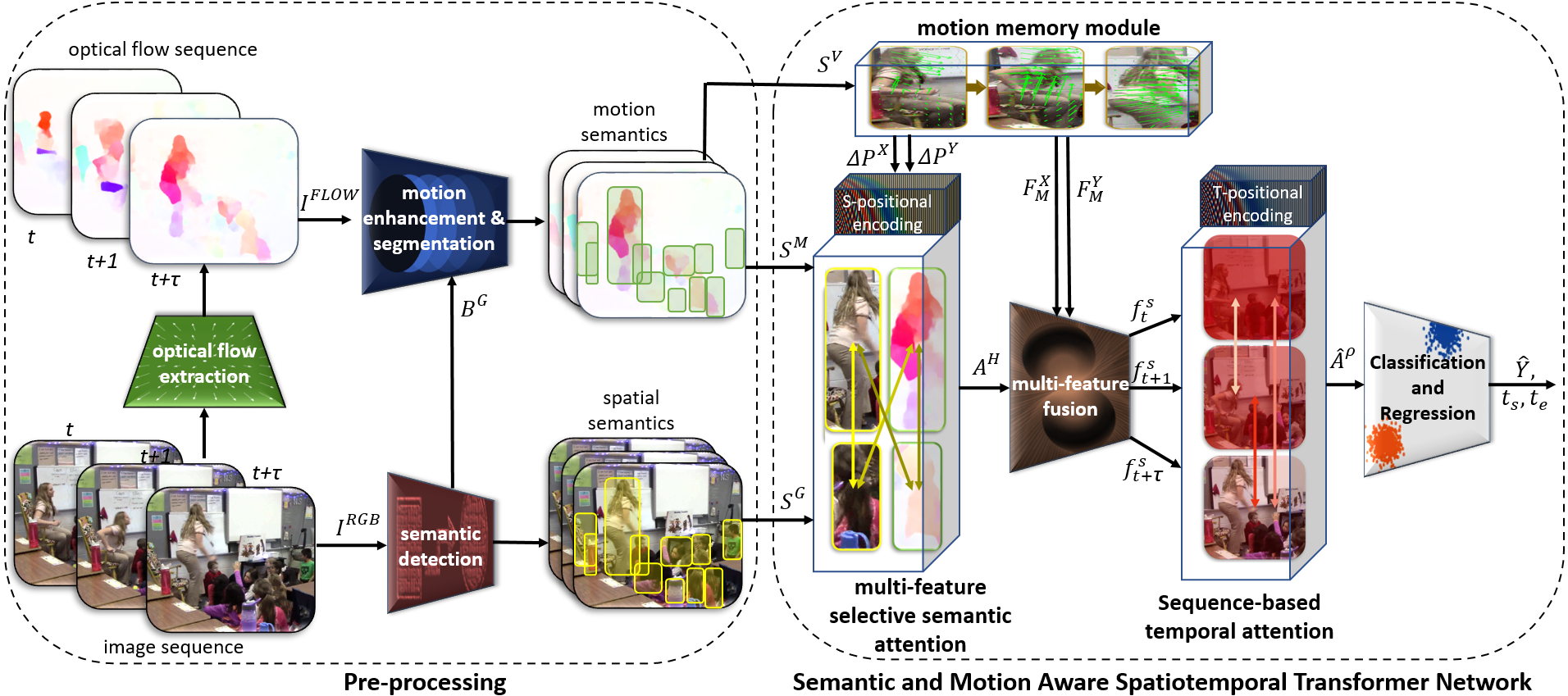}
	\caption{The pipeline for the proposed method includes the preprocessing stage and the transformer network. Given the sequence of RGB images, first, the spatial semantics and optical flow fields are extracted in the preprocessing stage. The motion enhancement and segmentation algorithm extracts the motion semantics that are invariant to camera movement. The multi-feature selective attention model captures the correlative patterns between spatial and motion semantics. The motion memory module updates the semantic positional encoding (S-positional encoding) of the transformer network and makes it semantically motion-aware. The multi-feature fusion combines the extracted features and directs them to the deep network. The sequence-based temporal attention model captures the heterogeneous temporal dependencies between different times that are then used to detect the action sequence in the classification and regression stage. 
	\label{Fig:pipeline}} 	
\end{figure*}

Fig. \ref{Fig:pipeline} shows the overview of the proposed pipeline for action detection.
Given a sequence of RGB frames, $I^{RGB}=\{I^{rgb}_t \in \mathbb{R}^{H \times W \times 3}, t=0,1,...,\tau\}$, the goal is to find the action class scores, $\hat{Y}$, and the start of the end of action, $t_s$ and $t_e$, respectively. Here, $\tau$ is the length of the action sequence; and $H$ and $W$ indicate the size of the full image (height and width).
The suggested pipeline includes two main stages: preprocessing (Section \ref{SEC:PRE}), in which the input data are prepared, and the spatiotemporal transformer network (Section \ref{SEC:SMATN}), which models and detects the action sequence.  

In the prepossessing stage, the spatial action semantics $S^G = \{Z^G, O^G\}$ are detected that includes the geometries of persons $Z^G=\{z^{g}_{i} \in \mathbb{R}^{h_i \times w_i \times 3}, i=0,1,...,N\}$ and objects $O^G=\{o^{g}_{i} \in \mathbb{R}^{h'_i \times w'_i \times 3}, i=0,1,...,N'\}$, where $N$, $N'$, $o^g$, and $z^g$ are the numbers of persons, number of objects, individual detected persons, and detected objects in an action frame, respectively. $h_i$ and $w_i$ indicate the sub-image size of the detected persons; and $h_i'$ and $w_i'$ shows the sub-image size of the detected objects. Note that the spatial action semantics represent the spatial information of persons and objects in RGB images.  We used \cite{carion2020ECCV} to detect the action semantics, including persons and objects.
Apart from the spatial features, motion features also are important properties of action semantics. So, to incorporate the motion features, the optical flow fields $I^{FLOW}=\{I^{flow}_t \in \mathbb{R}^{H \times W \times 3}, i=0,1,...,\tau\}$, are computed using \cite{kong2021ICRA}, a highly efficient optical flow extraction algorithm. The optical flow fields result from converting 2D optical flow motion vectors to three-channel images to enable computing the multi-feature attention between spatial and motion features (more details are in Section \ref{SEC:MFSSA}). Then, for each frame,  the motion semantics $S^M = \{Z^M, O^M\} \in I^{flow}_t$, including the motions of persons $Z^M=\{z^{m}_{i} \in \mathbb{R}^{h_i \times w_i \times 3}, i=0,1,...,N\}$ and objects $O^M=\{o^{m}_{i} \in \mathbb{R}^{h'_i \times w'_i \times 3}, i=0,1,...,N'\}$ are enhanced and segmented to make it invariant to camera movements (Section \ref{Sec:MH}).  The bounding boxes $B^G$, obtained from the semantic detection algorithm \cite{carion2020ECCV}, are used for motion segmentation. Here $B^G = \{B^Z, B^O\}$, where  $B^Z=\{b^{z}_{i} \in \mathbb{R}^4, i=0,1,...,N\}$ and $B^O=\{b^{o}_{i} \in \mathbb{R}^4, i=0,1,...,N'\}$; and $b^p$ and $b^o$ are the corresponding bounding boxes for each motion semantic.

The proposed spatiotemporal transformer network includes several modules to model the spatial and motion semantics obtained from the preprocessing step and detect the action sequence. The multi-feature selective semantic attention model (Section \ref{SEC:MFSSA}) captures the critical relationship between action semantics based on the correlations between their spatial and motion features. To handle the spatiotemporal variations in action semantics, the motion memory module (Section \ref{Sec:MMM}) utilizes the semantic motion vectors ($S^V$) and updates the semantic positional encoding of the transformer network using the 2D horizontal and vertical semantic motion memory offsets $\Delta p^x_i \in \Delta P^X$ and $\Delta p^y_i \in \Delta P^Y$, $i \in\{0,1, ,,,, N + N'\}$. Here, $S^V$ is computed from the optical flow motion vectors before conversion to images (more details are in Section \ref{Sec:MH}).

The motion memory module also outputs the 2D semantic motion memory features $F_M^X$ and $F_M^Y$.
The output of the multi-feature selective attention is the multi-head semantic attention, $A^H$, which represents the most informative selection of correlative patterns between spatial and motion semantics extracted by the heads of the transformer networks.      
The multi-feature fusion module (Section \ref{SEC:MFFM}) combines the motion memory and the multi-feature semantic features and directs the dataflow in multiple layers of the deep network.  The output of the multi-feature fusion module is the final set of semantic features, $f^s_t$, that are captured in different frames to form the sequence of final semantic , $X_s = \{f^s_t, t = 0,1,...,\tau\}$. Subsequently, $X_s$ is processed in the sequence-based temporal attention model (Section \ref{SEC:PTA}) to extract the heterogeneous temporal dependencies between different frames, $f^s_t$. The sequence-based temporal attention values, $\hat{A}$, proceed to the classification and regression stages to detect the action sequence. The implementation details of the proposed pipeline are in Section \ref{SEC:IMD}. We will discuss the aforementioned components of the proposed method thoroughly in the following sections. Our algorithm summary is shown in Table \ref{Tab:alg_sum}. The extended version of the algorithm summary is in the Supplementary Material.

\begin{table}[h!tbp]
	\centering
	\caption{Pipeline algorithm summary in a hierarchical order indicating each phase (in bold), the summary of each phase, and the inputs and outputs of each phase.} \label{Tab:alg_sum} 
\begin{tabular}{ccc}
\hline
Phase/Summary & Inputs & Outputs \\ \hline
\textbf{Semantic Detection} & \multirow{2}{*}{$I^{RGB}$} & \multirow{2}{*}{$S^G$}  \\ 
extracts action semantics & & \\ \hline 
\textbf{Optical Flow Extraction}  & \multirow{2}{*}{$I^{RGB}$} &  \multirow{2}{*}{$S^M$, $S^V$} \\
extracts optical flow & &  \\ \hline
\textbf{Motion Enhancement} & \multirow{2}{*}{$B^G$, $I^V$} & \multirow{2}{*}{$S^M$, $S^V$} \\
improves motion features &  &  \\ \hline
\textbf{Feature Embedding} & \multirow{2}{*}{$S^G$, $S^M$} & \multirow{2}{*}{$X^G$, $X^M$}  \\ 
extract spatiotemporal features  &  &  \\ \hline
\textbf{Motion Memory Module} & \multirow{2}{*}{$S^V$} & \multirow{2}{*}{$\Delta P^X$, $\Delta P^Y$}  \\ 
provides motion information  & & \\ \hline
\textbf{MA 2D Positional Encoding} & \multirow{2}{*}{$\Delta P^X$, $\Delta P^Y$} & \multirow{2}{*}{$P^X_A$, $P^Y_A$}  \\
for motion-aware transformer &  & \\ \hline
\textbf{MF Semantic Attention} & \multirow{2}{*}{$S^G$, $S^M$} & \multirow{2}{*}{$A^H$}\\
for correlations in multi-features &  & \\ \hline
\textbf{Multi-Feature Fusion} & \multirow{2}{*}{$A^H$} & \multirow{2}{*}{$X_s$} \\ 
combines features in layers & &  \\ \hline
\textbf{SB Temporal Attention} & \multirow{2}{*}{$X_s$} & \multirow{2}{*}{$\hat{A}$} \\ 
computes temporal relations &  &  \\ \hline
\textbf{Classification and Regression} & \multirow{2}{*}{$\hat{A}$} & \multirow{2}{*}{$\hat{Y}$, $t_s$, $t_e$}  \\
classifies actions and frames & & \\
\hline
\end{tabular}
\end{table}

\subsection{Preprocessing}
\label{SEC:PRE}
In the proposed pipeline, the preprocessing stage includes semantic detection, optical flow extraction, motion enhancement, and segmentation to prepare the inputs for the transformer network. We exploited \cite{carion2020ECCV}, a well-established state-of-the-art object/person detection algorithm, to extract spatial (RGB) action semantics, including persons and objects for each frame as $S^G = \{Z^G_t, O^G_t\}$. In this stage, the corresponding bounding boxes for persons and objects are  $B^G_t = \{B^Z_t, B^O_t\} \in \mathbb{R}^{(N + N') \times 4}$.

The optical flow fields are estimated from RGB images utilizing \cite{kong2021ICRA}, a highly efficient state-of-the-art optical flow estimation algorithm. Optical flow is a powerful and popular modality to represent motions in actions. However, optical flow is sensitive to camera movement, a common issue in videos captured in the wild. In other words, the camera movements can remarkably distort the motion information depicted in the optical flow fields. To address this problem, a semantic motion enhancement and segmentation algorithm is developed to use semantic motion vectors in the transformer network effectively.

\subsubsection{Semantic Motion Enhancement and Segmentation}
\label{Sec:MH}
 
Given the semantic bounding boxes $B^G = \{B^Z, B^O\}$ obtained from the semantic detection algorithm, the distorted optical flow fields $I^{flow}_t(x, y)$ (affected by camera movements) and the corresponding motion vectors, $I^{V}(u, v) \in \mathbb{R}^{H \times W \times 2}$, the goal of the semantic motion enhancement and segmentation algorithm is to extract the enhanced motion semantics including persons and objects $S^M = \{Z^M, O^M\}$; and the corresponding semantic motion vectors, $S^V = \{Z^V, O^V\}$, which are invariant to camera movements. Here, $u$ and $v$ are scalar units that define the motion displacement between the image pixels in the time $t$, as $(x^{(t)}, y^{(t)})$,  and the time $t+\omega$, as $(x^{(t+\omega)}, y^{(t+\omega)})$.  
The semantic motion enhancement and segmentation algorithm is designed based on the dominant motions of persons in action frames. Consequently, we consider persons as the foreground and the remaining portions of the frame as the background, affected mainly by camera movements \cite{korban2023PR}. The semantic motion enhancement and segmentation algorithm consists of two steps: motion modeling and motion restoration, which are explained in the Supplementary Material.

\subsection{Semantic and Motion-Aware Spatiotemporal Transformer Network}

\label{SEC:SMATN}

Before delivering the action semantics to the transformer network, they are embedded in the feature space. To do such, first, the outputs of the prepossessing step, including each spatial action semantic, $s^g_i \in S^G$, are resized as $\mathbb{R}^{h_i \times w_i \times 3} \rightarrow \mathbb{R}^{\hat{h} \times \hat{w} \times 2}$, and each motion semantic, $s^m_i \in S^{M}$ as  $\mathbb{R}^{h'_i \times w'_i \times 3} \rightarrow \mathbb{R}^{\hat{h} \times \hat{w} \times 2}$. Here, $\hat{h}$ and $\hat{w}$ represent the fixed image size.  So, the resized spatial and motion semantics are $\hat{S}^G$ and $\hat{S}^M$, respectively. 
Next, the action semantics are converted to spatial and motion semantic features as $X^G=\{X^G_Z, X^G_O\}$ and $X^M=\{X^M_Z, X^M_O\}$, respectively. This feature embedding is performed  using the convolutional layers, $Conv^G$, and $Conv^M$ as:

\begin{equation}
\label{EQ:EMD}
\begin{split}
     X^G = Conv^G(\hat{S}^G, W^G): \mathbb{R}^{\hat{N} \times \hat{h} \times \hat{w} \times 3} \rightarrow \mathbb{R}^{\hat{N} \times d_f}, \\
      X^M = Conv^M(\hat{S}^M, W^M): \mathbb{R}^{\hat{N} \times \hat{h} \times \hat{w} \times 3} \rightarrow \mathbb{R}^{\hat{N} \times d_f}, 
     \end{split}
\end{equation}

\noindent where $\hat{N} = N + N'$ is the total number of action semantics, $d_f$ is the size of feature embedding, and $W^G$ and $W^M$ are the kernel weights. The embedded feature set includes $X^G_Z$ and $X^G_O$, which are the spatial features of persons and objects; and $X^M_Z$, and $X^M_O$, which are the motion features of persons and objects, respectively. The aforementioned feature embedding is illustrated in Fig. \ref{Fig:MFSAP}.

\subsubsection{Motion-Aware 2D Positional Encoding.}
\label{Sec:MMM}
The transformer network has several advantages over traditional temporal networks because of the concurrent processing of inputs \cite{vaswani2017NIPS}  by using \emph{positional encoding} that adds the order information. As an integral part of the transformer network, positional encoding was initially designed for natural language processing, which was mainly a time-related problem.  The standard 1D temporal positional encoding \cite{vaswani2017NIPS} is illustrated as follows. 

\begin{equation}
\label{EQ:STA}
\begin{split}
P(p_t, 2n) = sin( \frac{p_t}{\psi^{\frac{2n}{d_m}}}), \\
P(p_t, 2n+1) = cos( \frac{p_t}{\psi^{\frac{2n}{d_m}}}),
\end{split}
\end{equation}

\noindent where $p_t$, $n$, and $d_m$ are the temporal order of the input, the dimension index of the positional embedding, and the semantic model size, respectively.  $\psi$ is a large integer as suggested by the original work \cite{vaswani2017NIPS} to accommodate high-dimensional embedding features. 

The positional encoding was later applied to spatial problems, such as object detection \cite{carion2020ECCV}, and spatiotemporal problems, such as action detection \cite{Liu2022TIP}, without effective adaptation to these areas.
Two main issues are (1) the 1D nature of the standard positional encoding, which leaves it less effective in handling 2D images; and (2) its incapability to handle spatiotemporal variations in spatiotemporal inputs such as action sequences. The proposed positional encoding has two characteristics that address the issues above. Firstly, the proposed positional encoding is ``2D'', making it more adept in dealing with 2D images. Secondly, and more importantly, the suggested positional encoding is motion-aware, which makes it more effective in handling spatiotemporal variations in action sequences.

\begin{figure*}[!htbp]
\renewcommand{\tabcolsep}{0.5pt}
	\centering 
		\begin{tabular}{ccc}
   \includegraphics[height=0.19 \textheight]{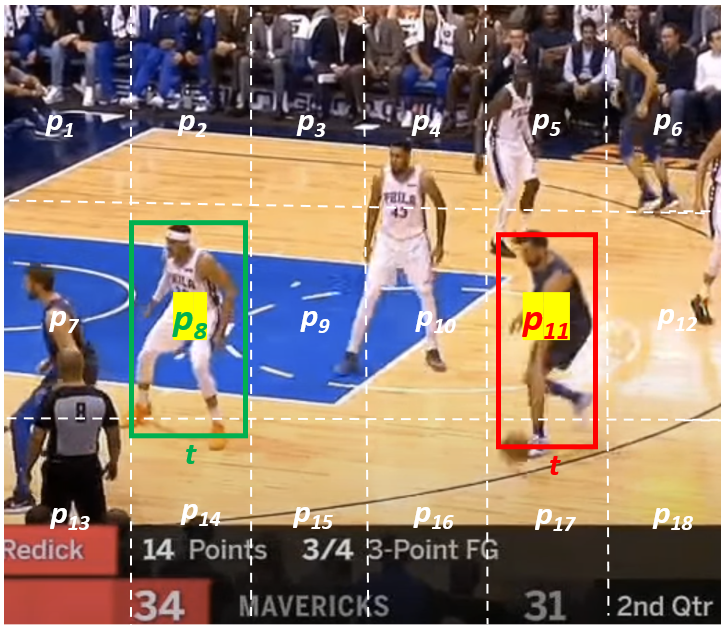} &
  \includegraphics[height=0.19 \textheight]{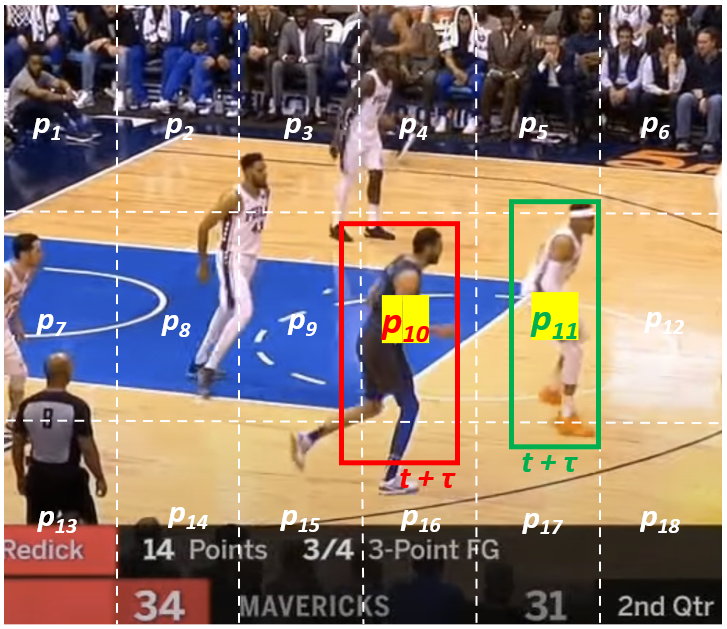} &
  \includegraphics[height=0.19 \textheight]{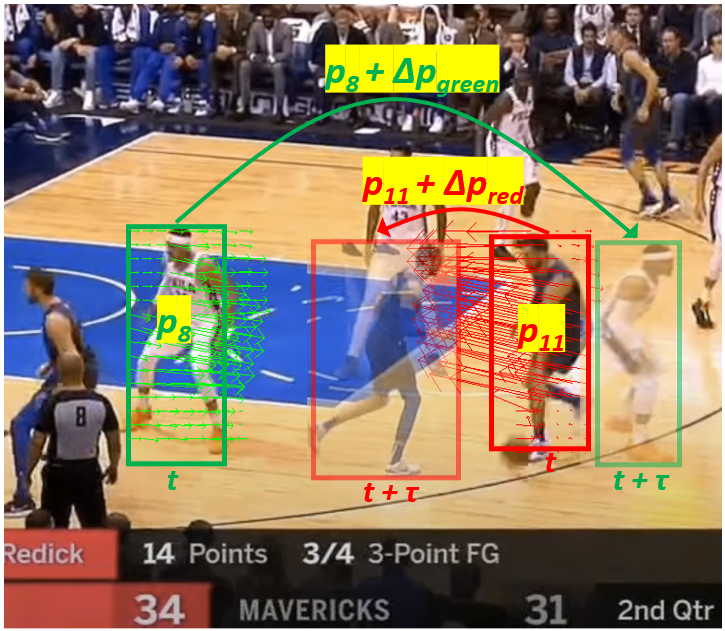} \\
  (a)  & (b) & (c) \\
  	\end{tabular}
	\caption{The motion-aware positional encoding, (c), compared to the standard one, (a) and (b), in dealing with spatiotemporal action semantic variations: two basketball players, red and green, change their positions from time $t$, (a) to  $t + \tau$, (b). So the red player ($p_{11}$ as the offensive player) switched his position to the green player (now $p_{11}$ the defender). The proposed motion-aware positional encoding, (c), can memorize the position changes of two basketball players using the motion memory offsets, $\Delta p_{green}$ and $\Delta p_{red}$. The green and red arrows show the motion vectors obtained from the optical flow fields. ~\label{Fig:2DPOS}}
\end{figure*}

Currently, the most common method for encoding spatial positions in images is to segment them into patches \cite{Dosovitskiy2020ICLR, khan2022ACS, han2021NIPS}. In this regard, the current methods treat 2D images as 1D inputs and assign the positional labels to each patch following the standard temporal positional encoding indicated in  (\ref{EQ:STA}). 
Nevertheless, the current methods employ fixed patches that cannot accommodate spatiotemporal variations in video frames. Fig. \ref{Fig:2DPOS} shows examples of such spatiotemporal variations and compares the proposed motion-aware positional encoding to the standard patch-based approach in dealing with this issue. In this example, the image is divided into 18 labeled between $p_1$ to $p_{18}$. Here, the red and green basketball players change their positions when moving from time $t$ (in Fig. \ref{Fig:2DPOS} (a)) to $t + \omega$ (in Fig. \ref{Fig:2DPOS} (b)). So, now, as a defender, the green player switches his position with the red player, on offense at position $p_{11}$. Consequently, when spatiotemporal changes occur in action semantics, the transformer network cannot model the semantic movements at different action frames when using  the standard patch-based positional encoding.  By contrast, the proposed motion-aware 2D positional encoding memorizes the position changes of each set of action semantics at different times and adaptively updates their positional information within the transformer network as shown in Fig. \ref{Fig:2DPOS} (c).  We later also numerically compare the proposed motion-aware 2D positional encoding to the standard one in Section \ref{SEC:ABL-PE} and Table \ref{Tab:MA-PE}. The motion-aware 2D positional encoding is formulated as follows:

\begin{equation}
\label{EQ:PU}
\begin{split}
    P^X_A(p^x_i, 2n) = sin( \frac{p^x_i +  \Delta p^x_i}{\psi^{\frac{2n}{d_m}}}), \\ P^X_A(p^x_i , 2n+1) = cos( \frac{p^x_i + \Delta p^x_i}{\psi^{\frac{2n}{d_m}}}),
        \end{split}
\end{equation}

\begin{equation}
\label{EQ:PV}
\begin{split}
    P^Y_A(p^y_i, 2n) = sin( \frac{p^y_i + \Delta p^y_i}{\psi^{\frac{2n}{d_m}}}), \\ P^Y_A(p^y_i, 2n+1) = cos( \frac{p^y_i + \Delta p^y_i}{\psi^{\frac{2n}{d_m}}}),
        \end{split}
\end{equation}

\noindent where $p^x_i$ and $p^y_i$ are initial horizontal and vertical positions for each action semantics $s_i \in \{S^G, S^M\}$, respectively that are obtained by the standard patch-base approach \cite{Dosovitskiy2020ICLR}.  $\Delta p^x_i \in \Delta P^X$, and $\Delta p^y_i \in \Delta P^Y$, $i=\{0, 1, ..., \hat{N}\}$ are horizontal and vertical semantic motion memory offsets for each action semantics. $\Delta P^X$ and $\Delta P^Y$ include the motion offsets for all the action semantics. $\omega$ is the duration of the motion. In this work, we define the index of positional embedding as $n \in \{0, 1, ..., d_f/2\}$.

$\Delta p^x_i$ and $\Delta p^y_i$ adaptively update the positions of action semantics according to their motion during the action sequence. A set of concatenated embedded 2D positional encoding $P_A = \{P^X_A, P^Y_A\}$ is delivered to the transformer network. 
To do such, the embedded positional encoding is added to the inputs as:

\begin{equation}
    \hat{X}^G = P_A + X^G, \quad \hat{X}^M = P_A + X^M,
\end{equation}

\noindent where $\hat{X}^G = \{\hat{X}^G_Z, \hat{X}^G_O\}$ and $\hat{X}^M = \{\hat{X}^M_Z, \hat{X}^M_O\}$ are positional encoded spatial and motion semantic features (for persons and objects), respectively. Two new components of the motion-aware positional encoding, $\Delta p^x_i$, and $\Delta p^y_i$ are computed using the motion memory module. Details of this computation are found below.

\textbf{Motion Memory Module.} 
Using a memory module has effectively modeled the varying temporal information \cite{korban2023PR3}.  The input to the motion memory module is the sequence of semantic motion vectors as $S^V_{seq} = \{ S^V_t, t = 0, 1, ..., \tau\}$. The outputs are (1) the semantic motion memory offsets,  $\Delta P^X$ and $\Delta P^Y$, that are used for our motion-aware 2D positional encoding; and (2) 2D semantic motion memory features, $F_M =\{ F_M^X, F_M^Y \}$ utilized in the multi-feature fusion stage to enrich the feature representation. The motion memory module memorizes the changes in semantic motion vectors using the motion memory network as follows:

\begin{equation}
\label{EQ:MMM}
\begin{split}
\text{update:} \quad z_t &= \sigma(W^z x_t + U^z h_{t- \hat{\omega}}), \\
\text{reset:} \quad r_t &= \sigma(W^r x_t + U^r h_{t- \hat{\omega}}), \\
\text{current:} \quad n_t &= \tanh(W^h x_t + r_t \odot U^h h_{t- \hat{\omega}}), \\
\text{output:} \quad h_t &= z_t \odot h_{t- \hat{\omega}} + (1 - z_t) \odot n_t,
\end{split}
\end{equation}

\noindent where $x_t$ is the input, $\odot$ represents the element-wise multiplication, and $\sigma$ is a sigmoid function.  $z_t$, $r_t$, $n_t$, and $h_t$ are the update gate, reset gate, current gate, and output, respectively. $W^z$, $W^r$, and $W^h \in \mathbb{R}^{\hat{N}\hat{h}\hat{w} \times d_g}$; and $U^z$, $U^r$, and $U^h \in \mathbb{R}^{d_g \times d_g}$ are parameter matrices, where $d_g$ is the state size. The above formulation is inspired by the learning steps of the gated recurrent unit \cite{ballakur2020ICCCS}.
Moreover, $\hat{\omega}$ is a dilated temporal value representing the time when significant motion exists in the action sequence based on a threshold value, $T_h$.
Given the inputs that are horizontal and vertical components of the semantic motion vectors, $S^V_X$, and $S^V_Y$, the memorized horizontal and vertical outputs of the networks are $h^X$ and $h^Y$, respectively. 
The motion memory networks memorize the movements of action semantics during action sequences. As a result, the motion memory networks update the positional encoding, making it motion-aware. The final outputs of the motion memory networks are computed by using an averaging pooling layer (\emph{AvPool}) as:

\begin{equation}
\label{EQ:MX}
\begin{split}
   \Delta P^X = AvPool(h^X) : \mathbb{R}^{\hat{N} \times d_g} \rightarrow \mathbb{R}^{\hat{N}}, \\
   \Delta P^Y = AvPool(h^Y): \mathbb{R}^{\hat{N} \times d_g} \rightarrow  \mathbb{R}^{\hat{N}},
    \end{split}
\end{equation}
    
The motion memory module also outputs the semantic motion memory features, $F_M = \{F_M^X, F_M^Y\}$, as:

\begin{equation}
\label{EQ:MFT}
\begin{split}
   F_M^X = Conv^{X}(h^X_Z, W^X), \quad F_M^X = Conv^{Y}(h^X_Z, W^Y).
    \end{split}
\end{equation}

Note that $Conv^X$ and $Conv^Y:\mathbb{R}^{N \times d_g} \rightarrow \mathbb{R}^{N \times d_f}$ and $W^X$ and $W^Y$ are the kernel weights.  
Only the part of the network outputs corresponding to persons as $h^X_Z$, and $h^X_Z \in \mathbb{R}^{N \times d_g}$ are selected in the feature selection. This is because persons' movements are more critical in action sequences than objects' movements. $F_M$ is used in the multi-feature fusion module to improve the feature representation of actions.

\subsubsection{Multi-feature Selective Semantic Attention}
\label{SEC:MFSSA}
Based on their spatial and motion features, many actions are characterized by spatiotemporal interactions between their semantics, including persons and objects.
Transformer networks are designed to capture the interactions between different inputs using an attention mechanism \cite{vaswani2017NIPS}. Therefore, we propose a multi-feature selective attention model to extract such interactions between action semantics. There are three differences between the proposed multi-feature attention and the standard ``cross-attention'' mechanism. First, the proposed attention is selective, meaning only informative queries are considered. Second, as opposed to the existing methods that include the whole frame/input, the suggested strategy focuses only on action semantics. Finally, four multi-feature attention types are introduced to enrich the action feature representation. 

\begin{figure*}[!htbp]
\renewcommand{\tabcolsep}{0.5pt}
	\centering 
		\begin{tabular}{cccc}
   \includegraphics[height=0.16\textheight]{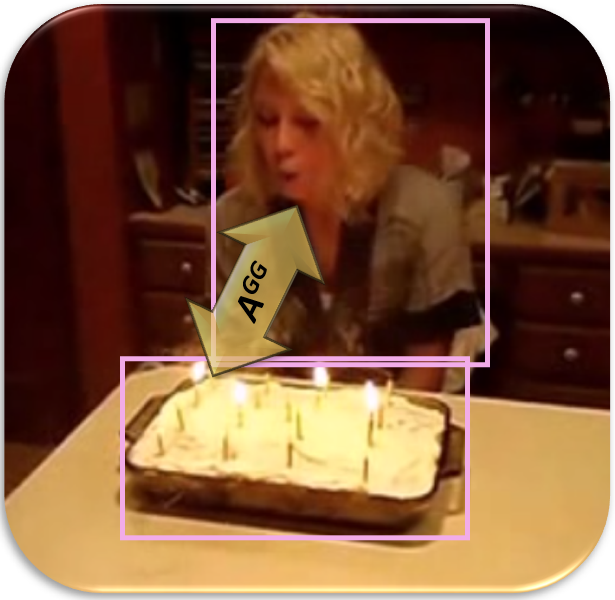} &
  \includegraphics[height=0.16\textheight]{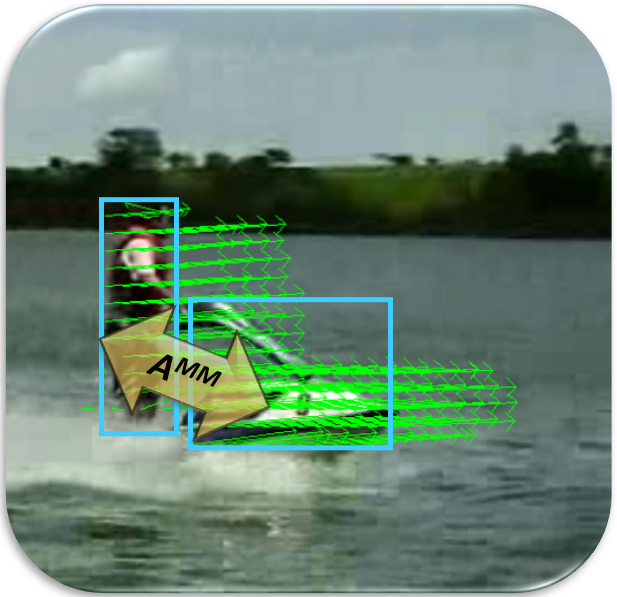} &
  \includegraphics[height=0.16\textheight]{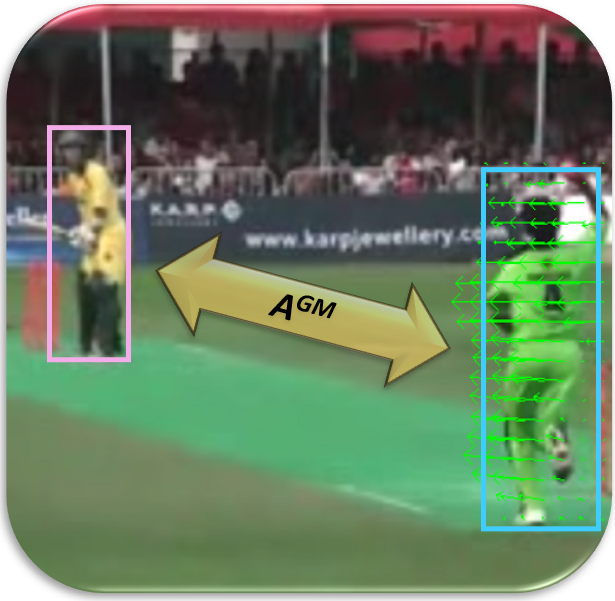} &
  \includegraphics[height=0.16\textheight]{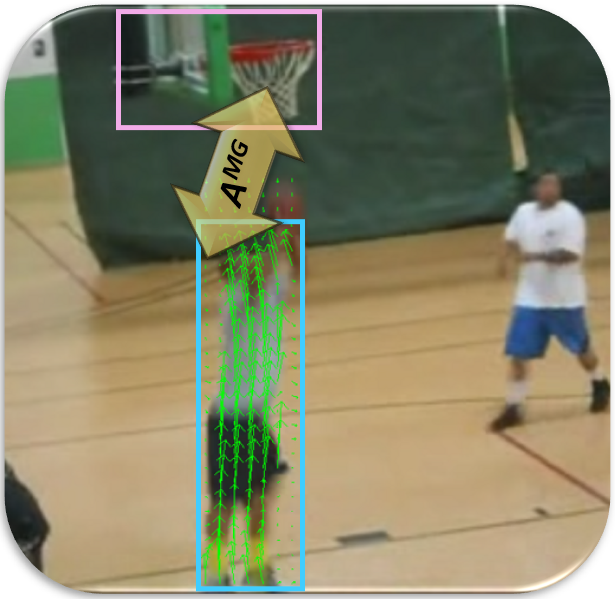} \\
  
  (a) & (b) & (c) & (d)\\
  	\end{tabular}
	\caption{Some examples of our multi-feature attention types that capture the spatiotemporal semantic interactions in action samples. (a): \emph{spatial-to-spatial attention} between the ``sitting person'' and the ``stationary cake'' ; (b): \emph{motion-to-motion attention} between the ``moving person'' and the ``moving jet ski'' ; (c): \emph{spatial-to-motion attention} between the ``stationary waiting player'' and the ``moving pitching player''; (d): \emph{motion-to-spatial attention} between the ``moving jumping player'' and the ``stationary net``. The green arrows show the motion vectors obtained from the optical flow fields.  The action samples are collected from the UCF101 dataset \cite{soomro2012ucf101}.  ~\label{Fig:MFE}}
\end{figure*}

 In our view, four types of multi-feature semantic interactions occur in human actions, which are illustrated in Fig. \ref{Fig:MFE}.  
The first scenario happens when all the interacting action semantics are stationary, such as the ``person" and ``cake/candle" in the action "blowing a candle" in Fig. \ref{Fig:MFE} (a). As a result, the spatial-to-spatial attention, $A^{GG}$, represents such a stationary-only interaction between action semantics.  In the second case, the entire action semantics are in motion, such as the "person" and the "jet ski" in the action "jet skiing" illustrated in  Fig. \ref{Fig:MFE} (b). Consequently, the motion-to-motion attention, $A^{MM}$, serves such motion-only interactions between action semantics. As for the third scenario, the spatial-to-motion attention, $A^{GM}$, provides the interaction between the moving semantics, the ``person pitching the ball", and the stationary one, the ``person waiting for the ball" in Fig. \ref{Fig:MFE} (c). Finally, the motion-to-spatial attention, $A^{MG}$, represents the interactions between moving and stationary semantics, in this example, basketball ``player" and ``net", respectively (shown in Fig. \ref{Fig:MFE} (d)).  Among the above multi-feature attentions types, $A^{GG}$ and $A^{MM}$ are intra-feature attention, while  $A^{GM}$ and $A^{MG}$ are inter-feature attention. 
In the aforementioned examples, the action semantics are highlighted by bounding boxes. The green motion vectors, which are obtained from optical flow fields, visualize the motion semantics.

\begin{figure*}[h!tbp]
	\centering
	\includegraphics[height=0.46\textwidth]{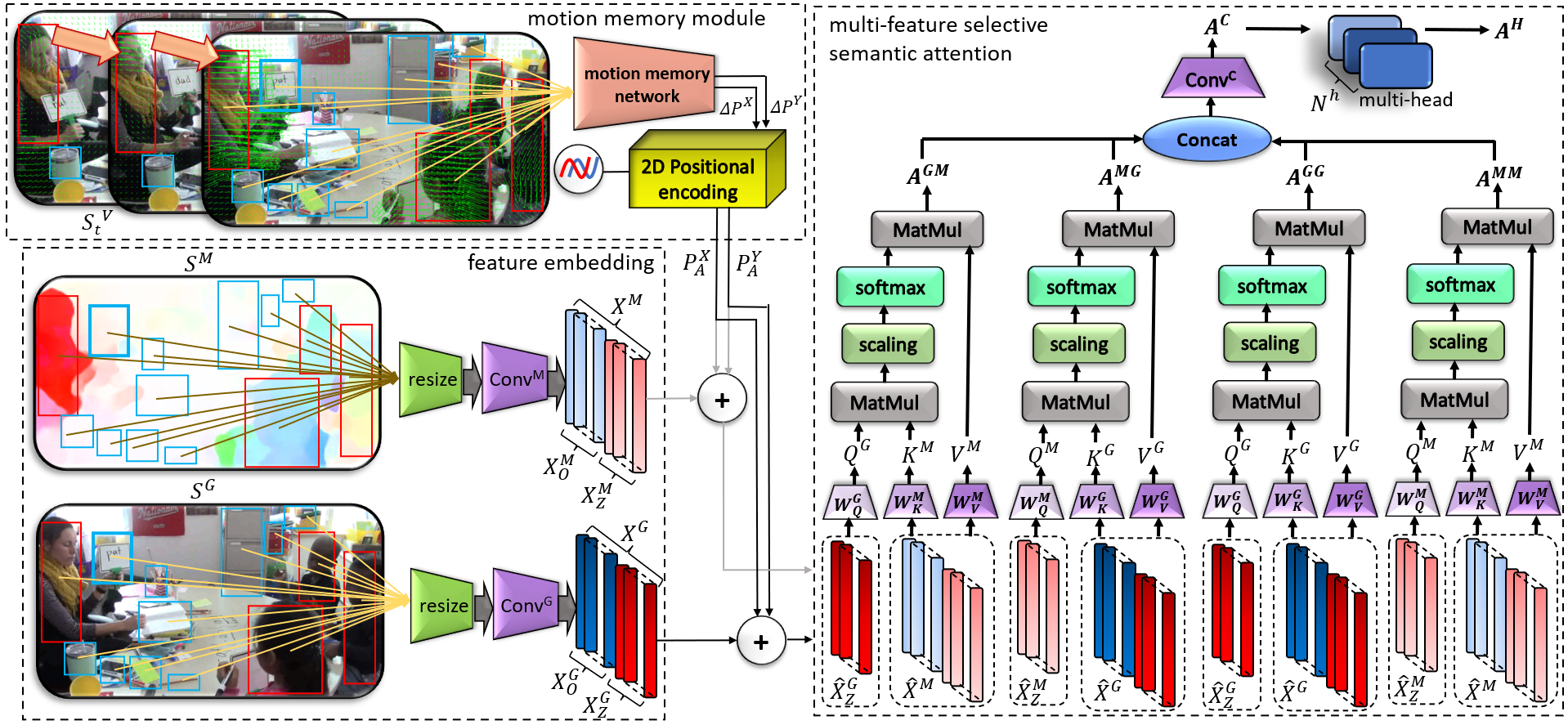}
	\caption{The proposed transformer network includes several modules to capture the multi-feature semantic features. The feature embedding converts the motion and spatial semantics to features. The motion memory module memorizes the semantic position changes and includes them as the motion-aware positional encoding in the semantic multi-feature extraction.  The multi-feature selective attention represents the correlations between persons with other persons and the most relevant objects. In this action example, ``teacher using an instructional tool'', these correlations represent the interactions between the ``teacher'' and the `students'', and the``teacher'' and relevant objects such as the ``handheld whiteboard.''    
	\label{Fig:MFSAP}} 	
\end{figure*}

The architecture of the multi-feature selective semantic attention model is illustrated in Fig. \ref{Fig:MFSAP}. The attention mechanism is to find the correlations between the \emph{query} ($Q$) and \emph{keys} ($K$) and then map them to \emph{values} ($V$). The four suggested attention types to represent the correlations between spatial and motion semantic features are shown in Fig. \ref{Fig:MFSAP} and are formulated as: 

\begin{equation}
\label{EQ:MATT}
\begin{split}
    A^{GM} = Softmax(\frac{Q^G(K^M)^T}{\sqrt{d_h}})V^M, \\
    A^{MG} = Softmax(\frac{Q^M(K^G)^T}{\sqrt{d_h}})V^G, \\
    A^{GG} = Softmax(\frac{Q^M(K^G)^T}{\sqrt{d_h}})V^G, \\
    A^{MM} = Softmax(\frac{Q^M(K^M)^T}{\sqrt{d_h}})V^M, \\
    \end{split}
\end{equation}

The multi-feature semantic queries, keys, and values are illustrated as follows:

\begin{equation}
\label{EQ:QKV}
\begin{split}
    Q^G = \hat{X}^G_ZW^G_Q, \;  K^G = \hat{X}^GW^G_K, \;  V^G = \hat{X}^GW^G_V, \\
        Q^M = \hat{X}^M_ZW^M_Q, \;  K^M = \hat{X}^MW^M_K, \;  V^M = \hat{X}^MW^M_V, 
    \end{split}
\end{equation}

\noindent where $W^G_Q, W^M_Q \in \mathbb{R}^{N \times d_f \times d_h}$, and $W^G_K, W^M_K, W^G_V, W^M_V \in \mathbb{R}^{\hat{N} \times d_f \times d_h}$ are projecting weights. $d_h = df / N^h$ is the size of attention head, $N^h$ is number of attention heads, and $T$ is a transpose operation.

The compound semantic attention head, $A^{C}$, is defined by combining the four multi-feature attention types and then applying a convolutional operation as:

\begin{equation}
\label{EQ:COM}
\begin{split}
    A^{C'} = Concat^C(A^{GM}, A^{MG}, A^{GG}, A^{MM}), \\
    A^{C} = Conv^C({A^{C'}, W^C}),
    \end{split}
\end{equation}

\noindent where $Conv^C: \mathbb{R}^{4 \times N\times d_h} \rightarrow \mathbb{R}^{N\times d_h}$, $Concat^C$ is a concatenation operator that stacks the inputs as  $Concat^C: 4 \times \mathbb{R}^{N\times d_h} \rightarrow \mathbb{R}^{4 \times N\times d_h}$, $W^C$ is the kernel weights, and $A^{C'}$ is the concatenated head. 
To extract features more effectively through different heads of the transformer network, the multi-head semantic attention,  $A^H \in \mathbb{R}^{N \times d_f}$ is computed by concatenating each of the $N^h$  compound semantic attention heads as:

\begin{equation}
\label{EQ:MH}
A^H = Concat^H(A^{C}_i, i=\{0,1,..., N^h\}),
\end{equation}

\noindent where $Concat^H: N^h \times \mathbb{R}^{N\times d_h} \rightarrow \mathbb{R}^{N\times N^h \times d_h} = \mathbb{R}^{N\times d_f}$.

\textbf{Selective Attention.}
As shown in Fig. \ref{Fig:MFSAP} and  (\ref{EQ:QKV}), in the proposed attention model, the queries and keys/values are defined differently. The queries are only persons, $\hat{X}_Z$, while the keys/values include both persons and objects, $\hat{X} = \{\hat{X}_Z, \hat{X}_O\}$. However, in the standard self-attention model, queries and keys/values are derived from the same source. We argue that the proposed selective attention model is more effective and efficient than the standard self-attention in modeling action semantics. In particular, the selective attention model computes the interactive correlations between persons-to-persons and persons-to-objects, which are essential to various actions. Therefore, we exclude the computation of correlations between objects-to-objects, which are often irrelevant to actions, thereby reducing the expressive power of action features. Specifically, among a large number of background objects in action videos, only a few are relevant to the action. An example is shown in Fig. \ref{Fig:MFSAP}, where the most relevant object is the ``Handheld whiteboard'' in the action ``teacher using an instructional tool''. By contrast, the ``teacher" is correlated with all the other persons, "students", in this activity. We will later numerically compare the performance of the selective attention model to the standard one in Section \ref{SEC:ABL-SA} and Table \ref{Tab:selfsel}. 
Moreover, the selective semantic attention is now represented in $\mathbb{R}^{N \times d_h}$, and comparing that expression to the standard self-attention space of $\mathbb{R}^{(N+N') \times d_h}$, leads to a reduced set of computations. Specifically, the computational cost of our selective semantic attention is $O(N \cdot (N + N') \cdot d_h)$, which is more efficient than a regular semantic self-attention with a computational cost that is $O((N + N')^2 \cdot d_h)$. The mathematical proof is provided in the supplementary material.

\subsubsection{Multi-feature Fusion}
\label{SEC:MFFM}
The proposed transformer network consists of multiple layers with a defined relationship between the successive layers $l-1$ and $l$, as illustrated as follows:

\begin{equation}
\begin{split}
\label{EQ:LAY1}
    \hat{B}_O^l = MFSSA(B_O^{l-1}) + B_O^{l-1}, \quad  l \in\{2,...,L\}, \\
    B_O^l = MLP(Norm(\hat{B}^{l})) + \hat{B}^{l}, \qquad l \in\{2,...,L\},
    \end{split}
\end{equation}

\noindent where $MFSSA$ is the multi-feature selective semantic attention, $B_O$ is the layer output, $\hat{B}_O$ is the intermediate layer output, $L$ is the number of layers, and $Norm$ is a normalization layer.

For the first layer, the relation between the inputs and output is shown as follows:

\begin{equation}
\label{EQ:MF}
    \hat{B^{1}_O} = MFSSA(\hat{X}^G, \hat{X}^M) + F^{M.A},
\end{equation}

\noindent where $F^{M.A} \in \mathbb{R}^{N \times d_h}$ is our motion-aware features defined as:

\begin{equation}
\label{EQ:MAF}
F^{M.A} = Conv^A(Concat^A(X^G_Z, X^M_Z, F^X_M, F^Y_M), W^A).
\end{equation}

\noindent where $Conv^A: \mathbb{R}^{4 \times N\times d_f} \rightarrow \mathbb{R}^{N\times d_f}$, $Concat^A: 4 \times \mathbb{R}^{N\times d_f} \rightarrow \mathbb{R}^{ 4 \times N\times d_f}$, and $W^A$ are the kernel weights. Here,  $MFSSA(\hat{X}^G, \hat{X}^M) = A^H$. To properly update the projecting weights for the keys and values in the layers $l \geq 2$ during the training, the selected inputs for the  multi-feature keys and values are converted to their original size of the first layer as $\mathbb{R}^{\hat{N}\times d_f}$. The final output of the multi-feature fusion for frame $t$ is the final semantic features, $f_t^s$ is illustrated as:  

\begin{equation}
\label{EQ:MOF}
f_t^s = Conv^F(B^L_O, W^O):  \mathbb{R}^{N\times d_f} \rightarrow \mathbb{R}^{d_f},
\end{equation}

\noindent where $W^O$ are the kernel weights. The sequence of $f_t^s$ is used as the multi-feature representation of semantics for each action frame in the sequence-based temporal attention model.

\subsubsection{Sequence-based Temporal Attention}
\label{SEC:PTA}
Given the sequence of final semantic features for different frames, $X_s = \{f^s_t, t=0,1,...,\tau\}$, the goal of the sequence-based temporal attention model, $\hat{A} \in \mathbb{R}^{\tau \times d_f}$ is to compute the temporal dependencies between semantic features in different frames $t\in \tau$, effectively. Temporal attention is capable of modeling long sequences without discriminating against older data, which makes it suitable for processing long, untrimmed action sequences. Nevertheless, the standard temporal attention was designed for natural language processing \cite{vaswani2017NIPS}. Accordingly, the standard temporal attention tends to assign the highest attention to words themselves and followed by words from similar categories. \cite{kalyan2021JBI}. However, such a homogeneous attention model is unsuitable for modeling heterogeneous temporal dependencies between action frames. There are many actions in which the temporal dependencies between distinctive frames, the so-called keyframes, provide essential information. For example, in the action sequence ``triple jump'' the temporal dependencies between the three distinctive steps ``hop'', ``step'', and ``jump'' define the meaning of the action \cite{korban2023PR2}. However, standard homogeneous temporal attention mainly focuses on the temporal dependencies between similar and often adjacent frames, discarding essential temporal dependencies between distinctive frames. To illustrate this phenomenon, we will show some examples in Section \ref{SEC:ABL-PTA}, and in Fig. \ref{Fig:PTATT}. 

On the other hand, the proposed sequence-based temporal attention model can compute the heterogeneous temporal dependencies between action frames without discriminating based on their similarity. It is because the sequence-based temporal attention model focuses on the temporal dependencies that are relevant for actions, rather than individual frames themselves. Some conceptual poof is provided in the supplementary material. We will later numerically compare the sequence-based temporal attention to the standard one in Section \ref{SEC:ABL-PTA} and Table \ref{Tab:tempatt}.

Some researchers suggested using a probabilistic approach to enhance the standard attention mechanism. \cite{gabbur2021NIPS} proposed exploiting a mixture model whose parameters are updated following the observed queries. \cite{nguyen2022ICML} presented a mixture of Gaussian keys as an effective replacement for redundant heads in a transformer network. These methods, however, are based on expectation maximization (EM), which has several issues, especially when used in a deep learning framework. This includes high computational cost, sensitivity to initialization, model complexity, and slow convergence \cite{abd2003ICS}. On the other hand, our sequence-based temporal attention is simple yet effective and does not require any complex and iterative optimization algorithms such as EM.

The standard temporal attention \cite{vaswani2017NIPS} is shown as: 

\begin{equation}
\label{EQ:TEMP}
    A(Q_{\Gamma}, K_{\Gamma}, V_{\Gamma}) = Softmax(\frac{Q_{\Gamma}K_{\Gamma}^T}{\sqrt{d_f}})V_{\Gamma},
\end{equation}

\noindent where the temporal queries, keys, and values are $Q_{\Gamma}$, $K_{\Gamma}$, and $V_{\Gamma} \in \mathbb{R}^{\tau \times d_f}$; and $\tau$ is the number of frames. $A^{corr}(Q_{\Gamma} , K_{\Gamma}) = Q_{\Gamma}K_{\Gamma}^T \in \mathbb{R}^{\tau \times \tau}$ is the temporal attention correlation matrix that indicates the frame-by-frame temporal dependencies. In the standard approach, $Q_{\Gamma} = (X_s + P)W_Q$, $K_{\Gamma} = (X_s+ P)W_K$, and $V_{\Gamma} = (X_s + P)W_V$, where $W_Q$, $W_K$, and $W_Q \in \mathbb{R}^{\tau \times d_f}$ are the temporal projecting weights. $P$ is the temporal positional encoding following  (\ref{EQ:STA}) that is added to the input sequence, $X_s$ to include the temporal order information. So, let's start with the definition of the standard temporal attention correlation matrix, $A^{corr}(Q_{\Gamma}, V_{\Gamma})$, that only focuses on the relationship between individual inputs (queries and keys):

\begin{equation}
\label{EQ:Temp1}
A^{corr}(Q_{\Gamma}, K_{\Gamma}) = (X_s+P)W_Q(X_s+P)^TW^T_K,
\end{equation}

The above can be rewritten as:

\begin{equation}
\label{EQ:Temp2}
\begin{split}
A^{corr}(Q_{\Gamma}, K_{\Gamma}) = \overbrace{X_sW_QX^TW^T_K}^{(a)} + \overbrace{PW_QP^TW^T_K}^{(b)}  + \\
\underbrace{X_sW_QP^TW^T_K}_{(c)} + \underbrace{PW_QX_s^TW^T_K}_{(d)},
\end{split}
\end{equation}

We argue that in the above, the expressions $(c)$ and $(d)$ do not contain useful information. This is because, during the learning process, they do not lead to meaningful gradient updates since they are the results of multiplying the input sequence and the position information. So, a more efficient temporal attention correlation matrix $\tilde{A}^{corr}(Q_{\Gamma}, K_{\Gamma})$ can be shown as follows: 

\begin{equation}
\label{EQ:Temp3}
\tilde{A}^{corr}(Q_{\Gamma}, K_{\Gamma}) = X_sW_QX^TW^T_K + {PW_QP^TW^T_K},
\end{equation}

The above can be rewritten as:

\begin{equation}
\label{EQ:Temp4}
\tilde{A}^{corr}(Q_{\Gamma}, K_{\Gamma}) = Q_{\Gamma}K_{\Gamma}^T + P_QP^T_K.
\end{equation}

 Consequently, the following defines the efficient temporal attention:
 
 \begin{equation}
 \label{EQ:tempeff}
     \tilde{A} = Softmax(\frac{\tilde{A}^{corr}}{\sqrt{d_f}})V_{\Gamma} 
 \end{equation}

Until now, the temporal dependencies between $Q_{\Gamma}$ and $K_{\Gamma}$  are calculated between individual frames without including their relation to the action sequence. To resolve this issue, the sequence-based temporal attention correlation matrix, $\hat{A}^{corr}$,  revises  (\ref{EQ:Temp4}) to provide the temporal dependencies between the inputs, $Q_{\Gamma}$ and $K_{\Gamma}$ with respect to the distribution of the action sequence, $X_s$, as shown as:

\begin{equation}
\begin{split}
\label{EQ:Tempour}
\hat{A}^{corr}(Q_{\Gamma}, K_{\Gamma}; X_s) = I_{\tau} - \\ (Q_{\Gamma} - K_{\Gamma})S_X^{-1}(Q_{\Gamma} - K_{\Gamma})^T/N_X  - \\
(P_Q - P_K)S_X^{-1}(P_Q - P_K)^T/N_P,
\end{split}
\end{equation}

\noindent where $S_X  \in \mathbb{R}^{d_f \times d_f}$ is the covariance matrix of the distribution $X_s$, and $I_{\tau} \in \mathbb{R}^{\tau \times \tau}$ is matrix of ones. $N_X$ and $N_P$ are normalizing terms.  The above also presents a more effective sequence-based temporal positional encoding than the original $P_Q$ and $P_K$.  (\ref{EQ:Tempour}) is based on the intuition of the Mahalanobis distance explained in Supplementary Material. Finally, the sequence-based temporal attention is computed as follows: 

\begin{equation}
\label{EQ:TEMF}
    \hat{A} = Softmax(\frac{\hat{A}^{corr}}{\sqrt{d_f}})V_{\Gamma},
\end{equation}

\subsubsection{Classification and Regression}
\label{SEC:CR}
The final stage of the proposed pipeline is classification and regression.  The multi-label classification layer includes two frame-level and sequence-level predictions shown as follows:

\begin{equation}
\begin{split}
\label{EQ:CLASS}
    \hat{Y}^S = Softmax(Conv^s(\hat{A}^P, W^s)), \\
    \hat{Y}^F = Softmax(Conv^f(\hat{A}^P, W^f)),
    \end{split}
\end{equation}

\noindent where  $\hat{Y}^S$, and $\hat{Y}^F$ are the sequence and frame prediction scores, respectively.  $Conv^s: \mathbb{R}^{\tau\times df} \rightarrow \mathbb{R}^{C_l}$, $Conv^f: \mathbb{R}^{\tau\times df} \rightarrow \mathbb{R}^{\tau \times C_l}$, and $W^s$ and $W^f$ are the kernel weights. Note that $C_l$ is the number of classes.  The set of class scores can be defined as $\hat{Y} = \{\hat{Y}^S,\hat{Y}^F\}$.
Finally, the regression layer calculates the start and end of the action sequence, $t_s$, and $t_e$, for a given class $c$ so that the selected time interval satisfies: $(t_s, t_e; c) = \underset{t}{\operatorname{argmax}} (\frac{1}{\tau}\sum_t(\hat{Y}^F_{t_s, t_e}) + \hat{Y}^S_{t_s, t_e})$. 

The loss function of the proposed network is shown as follows:

\begin{equation}
\label{LOSS}
    L = - \sum_{t=1}^{\tau} \sum_{c = 1}^{C_l} y_t^{(c)} log \hat{y}_t^{(c)} - \alpha\sum_{c = 1}^{C_l} Y^{(c)}log \hat{Y}^{(c)},
\end{equation}

\noindent where $y$ and $\hat{y}$ are the ground truth and predicted values for each frame $t$ of class $c$ at time $t$. $Y$ and $\hat{Y}$ are the ground truth and predicted values for each action sequence, respectively. Note that $\alpha$ is a loss adjustment parameter.

\section{Experimental Results}

\subsection{Implementation Details}
\label{SEC:IMD}

The implementation details of the proposed pipeline are summarized in Table 
\ref{Tab:impl}. In this table, all the hyperparameters mentioned in the paper are described.  The Model Zoo with the DETR architecture \cite{carion2020ECCV} and R101 backbone is used for object/person detection to balance efficiency and performance.  The object/person detection threshold is 0.5. The optical flow is extracted by the FastFlowNet \cite{kong2021ICRA} fine-tuned on FlyingThings3D \cite{mayer2016CVPR} with using $320 \times 448$ patches during data augmentation. 
The learnable semantic and temporal projecting weights $W^G_K$, $W^M_K$, $W^G_V$, $W^M_V$, $W_Q$, $W_K$, and $W_V$ are initialized with random values. For the EPIC-Kitchens \cite{damen2018ECCV}  dataset, only the classification layer is used. The network weights are initialized from the Kinetics pre-trained model. The focal loss \cite{lin2017ICCV} was used during the training. 

$S^V$ are required to update our motion-aware positional encoding, which is also 2D.  On the other hand, $S^M$ helps calculate multi-feature attention between motion and spatial features that are also derived from images. Moreover, the motion feature embedding in our pipeline is computed by a convolutional layer requiring images as the inputs. 

All the experiments are conducted using PyTorch 1.7 on a server PC with dual Nvidia RTX 3090 GPUs (24GB VRAM), AMD Ryzen Threadripper 3990X 64-Core Processor, and 256GB of RAM.

\begin{table}[h!tbp]
	\centering
	\caption{Implementation details of the proposed pipeline with corresponding sections.} \label{Tab:impl} 
\begin{tabular}{ccc}
\hline
Parameter & Value & Section\\ \hline
Semantic detection confidence threshold & 0.5 &  \ref{Sec:MH} \\
Number of GMM distributions ($K$)   & 16 &  \ref{Sec:MH} \\ 
Number of action semantic ($\hat{N}$) & 10 &   \ref{SEC:SMATN} \\
Semantic image size ($\hat{h} \times \hat{w}$) & $128 \times 128$ & \ref{SEC:SMATN} \\
Size of feature embedding ($d_f$) & 2048 &  \ref{SEC:SMATN} \\
Positional encoding integer ($\psi$) & $1e^{4}$ & \ref{Sec:MMM} \\
Duration of motion  ($\tau$) & 32 &  \ref{Sec:MMM} \\
Size of motion network state ($d_g$) & 1024 & \ref{Sec:MMM} \\
Motion memory module threshold  ($T_h$) & 2.35 &  \ref{Sec:MMM} \\
Number of attention heads ($N^h$) & 4 & \ref{SEC:MFSSA} \\
Size of attention head ($d_h$) & 512 & \ref{SEC:MFSSA} \\
Number of multi-feature layers  ($L$) & 6 & \ref{SEC:MFFM} \\
Input normalizing term $(N_X)$ & $1e^{3}$ & \ref{SEC:PTA} \\
Position normalizing term $(N_P)$ & $2e^{3}$ & \ref{SEC:PTA} \\
Loss adjustment parameter ($\alpha$) & 2.4 & \ref{SEC:CR} \\
Optimizer  & SGD &  \ref{SEC:CR} \\
Number of training epochs & 100 & \ref{SEC:CR} \\
Number of videos per batch & 16 & \ref{SEC:CR} \\
Learning rate & $1e^{-4}$ &   \ref{SEC:CR} \\
Learning rate decay (every 30 epochs) & 0.1 &   \ref{SEC:CR} \\
Weight decay & $1e^{-6}$ &  \ref{SEC:CR} \\
Multi-label classification threshold & 0.5 & \ref{SEC:CR} \\
\hline
\end{tabular}
\end{table}

\subsection{Datasets}

The proposed pipeline is evaluated using four public spatiotemporal action datasets, AVA (version 2.1 and 2.2) \cite{gu2018CVPR}, UCF101-24 \cite{soomro2012ucf101}, and EPIC-Kitchens \cite{damen2018ECCV}. Short descriptions of these datasets are given in the following. 

\textbf{AVA} dataset comprises 15-minute video segments of 80 action classes. The AVA dataset is suitable for evaluating our proposed pipeline as it covers a notable variety of interactions between action semantics, including person-to-person and person-to-object interactions. 
Both AVA versions (v2.1 and v2.2) include 235 videos for training and 64 videos for validation.  The annotations for bounding boxes and class labels are provided for every second of the video segments.  AVA 2.2 revises the label annotation by adding more frame-level labels. 
Following the instruction of the benchmark, \cite{gu2018CVPR} and the previous studies \cite{tang2020ECCV, wu2019CVPR, pan2021CVPR}, we used 60 action classes with the measurement metrics of \emph{mean Average Precision (mAP)} with an IoU threshold of 0.5.  

\textbf{UCF101-24} is a subset of the UCF101 dataset \cite{soomro2012ucf101} that includes 3'207 untrimmed videos from 24 action classes. As a part of THUMOS Challenge 2015, this dataset includes the annotation for bounding boxes and frames. It covers a variety of spatial and temporal action instances in each video, which is well-suited for our experiments. An mAP with an IoU threshold of 0.5 is used for evaluation on the first split of the data following the previous work \cite{tang2020ECCV, pan2021CVPR}.

\textbf{EPIC-Kitchens} comprises 55 hours of daily activities, primarily focusing on the interaction between persons and objects. This dataset features the most diverse objects among spatiotemporal action benchmarks, making it suitable to test the effectiveness of the proposed pipeline in modeling action semantics. Following a standard setting \cite{tang2020ECCV, baradel2018ECCV}, we used 22’675 videos for training and 5’886 videos for testing.  According to the literature \cite{girdhar2022omnivore_cvpr, wu2022CVPR, herzig2022CVPR}, the \emph{top-1} accuracy metric is adopted to evaluate the suggested method on the EPIC-Kitchens dataset on ``action'', ``noun'', and ``verb'' classes. A \emph{top-1} accuracy is obtained by comparing the highest predicted results with the ground truth.

\subsection{Comparative Results}
The comparative results of four public benchmarks are as follows.

\subsubsection{Results on AVA 2.2}

Table \ref{Tab:compareAVA22} illustrates the comparison between our method, SMAST, and the state-of-the-art approaches on the AVA 2.2 dataset.  In \cite{tong2022NIPS}, a video-masked autoencoder using pre-trained models achieved the highest transfer learning performance when using Kinetic-700. 
Overall, our proposed pipeline (SMAST) outperforms the state-of-the-art approaches on the AVA 2.2 dataset. Specifically, with the mAP@0.5 of \textbf{40.2\%}, the suggested method surpasses the best current benchmark \cite{tong2022NIPS} by 0.9. 

\begin{table}[h!tbp]
	\centering
	\caption{Comparison of our proposed method (SMAST) with the state-of-the-art strategies on the AVA 2.2 dataset.} \label{Tab:compareAVA22} 
\begin{tabular}{cccc}
\hline
Team & Method & Pub/Year &  mAP@0.5 (\%) \\ \hline
Tang et al. \cite{tang2020ECCV} & AIA & ECCV 2020 & 34.4  \\
Feichten al. \cite{feichten2021CVPR} & MoCo & CVPR 2021 & 20.3 \\
Fan et al. \cite{fan2021ICCV} & MVT & ICCV 2021 & 27.3 \\
Feichten al. \cite{feichtenhofer2022CVPR} & X3D & CVPR 2022 & 27.4 \\
Chen et al. \cite{chen2021ICCV} & WOO & ICCV 2021 & 28.3 \\
Wu et al. \cite{wu2021CVPR} & TLVU & CVPR 2021 & 31.0 \\
Pan et al. \cite{pan2021CVPR} & ACARNet & CVPR 2021 & 31.7  \\
Liu et al. \cite{herzig2022CVPR} & ORVT & CVPR 2022 & 26.6\\
Zhao et al. \cite{zhao2022CVPR} & Tuber & CVPR 2022 & 33.6 \\
Wu et al. \cite{wu2022CVPR} & MeMViT & CVPR 2022 & 35.4  \\
Wei et al. \cite{wei2022CVPR} & MaskFeat & CVPR 2022 & 38.8 \\
Tong et al. \cite{tong2022NIPS} & VideoMAE & NIPS 2022 & 39.3 \\
Faure et al. \cite{faure2023WCACV} & HIT & WCACV 2023 & 32.6  \\
\textbf{Korban et al.} & \textbf{SMAST} & - - - - -  & \textbf{40.2} \\ \hline
\end{tabular}
\end{table}

\subsubsection{Results on AVA 2.1}

Table \ref{Tab:compareAVA21} compares The proposed method, SMAST, to the state-of-the-art strategies on the AVA 2.1 dataset. 
In conclusion, the proposed SMAST surpassed the state-of-the-art methods previously used on the AVA 2.1 dataset with the mAP@0.5 of \textbf{33.1\%}, exceeding the highest existing benchmark \cite{zhao2022CVPR} by 1.1\%.

\begin{table}[h!tbp]
	\centering
	\caption{Comparison of our suggested method (SMAST) with the state-of-the-art approaches on the AVA 2.1 dataset.} \label{Tab:compareAVA21} 
\begin{tabular}{cccc}
\hline
Team & Method & Pub/Year &  mAP@0.5 (\%) \\ \hline
Sun et al. \cite{sun2018ECCV} & ACRN & ECCV 2018 & 17.4 \\ 
Girdhar et al. \cite{girdhar2019CVPR} & VT & CVPR 2019 & 27.6 \\ 
Wu et al. \cite{wu2019CVPR} & LTFB & CVPR 2019 & 27.7 \\
Stroud et al. \cite{stroud2020CVPR} & D3D & CVPR 2020 & 23.0 \\ 
Tang et al. \cite{tang2020ECCV} & AIA & ECCV 2020 & 31.2  \\ 
Wu et al. \cite{wu2021CVPR} & TLVU & CVPR 2021 & 27.8 \\
Chen et al. \cite{chen2021ICCV} & WOO & ICCV 2021 & 28.0 \\ 
Pan et al. \cite{pan2021CVPR} & ACARNet & CVPR 2021 & 30.0  \\ 
Shah et al. \cite{shah2022CVPR} & PGA & CVPR 2022 & 28.4 \\ 
Zhao et al. \cite{zhao2022CVPR} & TubeR & CVPR 2022 & 32.0 \\
\textbf{Korban et al.} & \textbf{SMAST} & - - - - -  & \textbf{33.1} \\ \hline
\end{tabular}
\end{table}

\subsubsection{Results on UCF101-24}

Table \ref{Tab:compareUCF} illustrates the comparative results on the UCF101-24 benchmark. 
SMAST, outperformed the other strategies on the UCF101-24 benchmark with the mAP@0.5 of \textbf{85.5\%}, exceeding the highest performance \cite{faure2023WCACV} by 0.7.

\begin{table}[h!tbp]
	\centering
	\caption{Comparison of our pipeline (SMAST) with the state-of-the-art methods on the UCF101-24 dataset.} \label{Tab:compareUCF} 
\begin{tabular}{cccc}
\hline
Team & Method & Pub/Year &  mAP@0.5 (\%) \\ \hline
Song et al. \cite{song2019CVPR} & Tacnet & CVPR 2019 & 72.1 \\
Pramono et al. \cite{pramono2019CVPR} & HSAN & CVPR 2019 & 73.7 \\
Yang et al. \cite{yang2019CVPR} & STEP & CVPR 2019 & 75.0 \\
Zhang et al. \cite{zhang2019CVPR} & ASM & CVPR 2019 & 77.9 \\
Tang et al. \cite{tang2020ECCV} & AIA & ECCV 2020 & 78.8  \\  
Li et al. \cite{li2020ECCV} & MOC & ECCV 2020 & 78.0 \\
Su et al. \cite{su2020PAMI} & PCSC & PAMI 2020 & 79.2 \\
Liu et al \cite{liu2021ACDnet} & ACDnet & PRL  2021 & 70.9 \\
Pan et al. \cite{pan2021CVPR} & ACARNet & CVPR 2021 & 84.3  \\ 
Kumar et al. \cite{kumar2022CVPR} & ESSL & CVPR 2022 & 69.9 \\
Li et al. \cite{li2022SIVP} & DSRM & SIVP 2022 & 81.2  \\
Zhao et al. \cite{zhao2022CVPR} & TubeR & CVPR 2022 & 81.3\\
Faure et al. \cite{faure2023WCACV} & HIT & WCACV 2023 & 84.8  \\
\textbf{Korban et al.} & \textbf{SMAST} & - - - - -  & \textbf{85.5} \\ \hline
\end{tabular}
\end{table}

\subsubsection{Results on EPIC-Kitchens}
\ref{Tab:comparekit} compares the proposed pipeline, SMAST, and state-of-the-art methods on the EPIC-Kitchens benchmark. Our proposed method outperformed the existing approaches with the top-1 accuracy for ``action'', ''verb'', and ``noun'' of \textbf{50.9\%}, \textbf{70.1\%}, and  \textbf{64.8\%}, respectively.

\begin{table}[h!tbp]
	\centering
	\caption{Comparative results on the EPIC-Kitchens dataset showing top-1 accuracy (\%) for different classes of action, noun, and verb.} \label{Tab:comparekit} 
\begin{tabular*}{\columnwidth}{@{\extracolsep{\fill}}cccccc}
\hline
Team & Method & Pub/Year &  action & verb & noun\\ \hline
Tang et al. \cite{tang2020ECCV} & AIA & ECCV 2020 & 27.7 & 60.0 & 37.2 \\
Nagrani et al. \cite{nagrani2021NIPS} & MBT & NIPS 2021 & 43.4 & 64.8 & 58.0 \\
Arnab et al. \cite{arnab2021vivit_iccv} & Vivit & ICCV 2021 & 44.0 & 66.4 & 56.8 \\
Patrick et al. \cite{patrick2021NIPS} & Mformer & NIPS 2021 & 44.5 & 67.0 & 58.5 \\
Kondrat et al. \cite{kondratyuk2021movinets_cvpr} & Movinets & CVPR 2021 & 47.7 & 72.2& 57.3 \\
Liu et al. \cite{herzig2022CVPR} & ORVT & CVPR 2022 & 45.7 & 68.4 & 58.7 \\
Wu et al. \cite{wu2022CVPR} & MeMViT & CVPR 2022 & 48.4 & 71.4 & 60.3 \\
Girdhar et al. \cite{girdhar2022omnivore_cvpr} &  Omnivore & CVPR 2022  & 49.9 & 69.5 & 61.7 \\
Yan et al. \cite{yan2022CVPR} & MVT & CVPR 2022 & 50.5	& 69.9  & 63.9 \\
\textbf{Korban et al.} & \textbf{SMAST} & - - - - -  & \textbf{50.9} & \textbf{70.1}  & \textbf{64.8} \\ \hline
\end{tabular*}
\end{table}

\subsection{Ablation Study}
\label{SEC:ABL}
In this section, we present an ablation study that evaluates the impact of each module in the proposed action detection solution. 

\subsubsection{Motion-Aware 2D Positional Encoding and Motion Enhancement}
\label{SEC:ABL-PE}
Table \ref{Tab:MA-PE} indicates the impact of the proposed motion-aware (MA) 2D positional encoding and the proposed motion enhancement algorithm on the overall action detection performance.  Our findings show that the MA 2D positional encoding is ineffective without the motion enhancement algorithm. Specifically, it only boosted the overall performance by 0.1 \% (from 30.9 \% to 31.0\%). The reason is that, without motion enhancement, the motion awareness of 2D positional encoding is impacted by the camera movement, a common issue in action videos captured in the wild. Notably, the camera movement causes incorrect extraction of motion vectors, resulting in imprecise calculations of the motion memory offsets, $\Delta P^X$ and $\Delta P^Y$, in the 2D positional encoding. In contrast, when the motion enhancement algorithm is used, the MA 2D positional encoding increases the overall performance by 2.2\% (from 30.9\% to 33.1\%). The 1D positional encoding is computed following \cite{Dosovitskiy2020ICLR}.

\begin{table}[h!tbp]
	\centering
	\caption{The impact of the motion-aware (MA) 2D positional encoding and motion enhancement algorithm on the overall action detection performance. The evaluation is conducted on the AVA 2.1 benchmark.} \label{Tab:MA-PE} 
\begin{tabular*}{\columnwidth}{@{\extracolsep{\fill}}cc}
\hline
Scenario & mAP@0.5  \\ \hline
Standard 1D positional encoding & 30.9 \\
MA 2D positional encoding (\textbf{no} motion enhancement) &  31.0 \\
\textbf{MA 2D positional encoding} (\textbf{with} motion enhancement) &  \textbf{33.1}\\
\hline
\end{tabular*}
\end{table}

\subsubsection{Multi-Feature Semantic Attention}
Table \ref{Tab:MFA} illustrates the impact of various multi-feature attention types and their combinations on the overall action detection performance (tested on the AVA 2.1 dataset). When a single attention type is used, the highest performance is achieved by the intra-feature attention types, including $A^{GG}$ with a mAP@0.5 of 26.5\% followed by $A^{MM}$, with a mAP@0.5 of 25.6\%. The inter-feature attention types, $A^{MG}$, and $A^{GM}$, alone did not result in competitive performance. However, combined with $A^{GG}$ and $A^{MM}$, the inter-feature attention types, $A^{MG}$, and $A^{GM}$, boosted the overall action detection performance. The results show that although many actions can still be effectively modeled independently, many depend on the interaction between spatial and motion features. The highest performance is obtained when all the multi-feature attention types, $A^{GG}$, $A^{MM}$, $A^{GM}$, and $A^{MG}$ are used, yielding a mAP@0.5 of 33.1\%.

\begin{table}[h!tbp]
	\centering
	\caption{The impact of different multi-feature (MF) attention types, $A^{GG}$ (spatial-to-spatial), $A^{MM}$ (motion-to-motion), $A^{GM}$ (spatial-to-motion), and $A^{MG}$ (motion-to-spatial) on the overall action detection performance. Different attention types represent standard one-stream and two-stream baselines and our single, double, combined, and full MF attentions. } \label{Tab:MFA}
 \begin{tabular}{cc}
\hline
Attention type & mAP@0.5 (\%)  \\ \hline
$A^{GG}$ (one-stream RGB baseline) & 26.5\\
$A^{MM}$ (one-stream FLLOW baseline) & 25.6 \\
$A^{GM}$ (single MF attention) & 24.7\\
$A^{MG}$ (single MF attention)& 24.4\\
$A^{GG}$+ $A^{MM} $ (two-stream RGB+FLOW baseline)& 28.0\\
$A^{GM}$+ $A^{MG}$ (double MF attention) & 27.1\\
$A^{GG}$+ $A^{MM}$+ $A^{GM}$ (combined MF attention) & 31.8\\
$A^{GG}$+ $A^{MM}$+ $A^{MG}$ (combined MF attention) & 30.9\\
$\bm{A^{GG}+ A^{MM}+ A^{GM}+ A^{MG}}$ (full model) & \textbf{33.1}\\
\hline
\end{tabular}
\end{table}

\subsubsection{Selective Attention}
\label{SEC:ABL-SA}

Table \ref{Tab:selfsel} compares various forms of self-attention and the proposed selective attention model characterized by two types of action semantics, persons ($Z$) and objects ($O$).  The lowest performance is due to the self-attention mechanism that defined queries and keys as objects (mAP@0.5 of 24.3\%). In contrast, the maximum performance among the self-attention types is achieved with the queries and keys specified as persons + objects ($Z + O$) with a mAP@0.5 of 31.8\%. Overall, the proposed selective attention model with the queries defined as persons ($Z$) and the keys as persons + objects ($Z + O$) yields a mAP@0.5 of 33.1\%. 

\begin{table}[h!tbp]
	\centering
	\caption{The comparison between different self-attention types and the proposed selective attention model based on two action semantics, persons ($Z$), and objects ($O$). Here the self-attention follows the standard attention mechanism in the baseline \cite{vaswani2017NIPS}. } \label{Tab:selfsel}
\begin{tabular*}{\columnwidth}{@{\extracolsep{\fill}}ccc}
\hline
Attention category & Inputs (Semantics) & mAP@0.5 (\%)  \\ \hline
Self-attention  & Query ($O$) , Key ($O$) & 24.3 \\
Self-attention  & Query ($Z$) , Key ($Z$) & 27.5 \\
Self-attention  & Query ($Z + O$) , Key ($Z + O$) & 31.8 \\
\textbf{Selective attention} &  \textbf{Query} ($\bm{Z}$) , \textbf{Key} ($\bm{Z + O}$) & \textbf{33.1} \\
\hline
\end{tabular*}
\end{table}

\subsubsection{Sequence-based Temporal Attention}
\label{SEC:ABL-PTA}
Table \ref{Tab:tempatt} compares the suggested sequence-based temporal attention with the efficient and standard ones on the AVA 2.1 dataset. In this table, the standard temporal positional encoding (TPE) was presented in  (\ref{EQ:Temp1}), the efficient temporal attention (ETPE) in  (\ref{EQ:Temp4}), and the sequence-based temporal positional encoding (PTPE) in  (\ref{EQ:Tempour}). Moreover, $A$, $\tilde{A}$, and $\hat{A}$ were shown in  (\ref{EQ:TEMP}),  (\ref{EQ:tempeff}), and  (\ref{EQ:TEMF}), respectively. 

The results illustrated that the efficient temporal attention ($\tilde{A}$) + ETPE  with a mAP@0.5 of 30.9\% slightly performed better than the standard temporal attention ($A$) + TPE with a mAP@0.5 of 30.4\%. Using sequence-based models for both temporal attention and positional encoding, $\hat{A}$ + PTPE, yielded the maximum performance with a mAP@0.5 of 33.1. 

\begin{table}[h!tbp]
	\centering
\caption{The comparison between standard temporal attention, efficient temporal attention, and the proposed sequence-based temporal attention.  Different types of positional encodings are the standard temporal positional encoding (TPE), efficient temporal positional encoding (ETPE),  and the proposed sequence-based temporal positional encoding (PTPE).} \label{Tab:tempatt}
\begin{tabular*}{\columnwidth}{@{\extracolsep{\fill}}cc}
\hline
Modules & mAP@0.5 (\%)  \\ \hline
Standard temporal attention ($A$) + TPE  (baseline \cite{vaswani2017NIPS}) & 30.4 \\ 
Efficient temporal attention ($\tilde{A}$)  + ETPE & 30.9 \\ 
sequence-based temporal attention ($\hat{A}$) + ETPE & 32.5 \\
\textbf{sequence-based temporal attention} ($\bm{\hat{A}}$) + \textbf{PTPE}  & \textbf{33.1}  \\
\hline
\end{tabular*}
\end{table}

\begin{figure*}[h!tbp]
	\centering
	\includegraphics[height=0.36\textwidth]{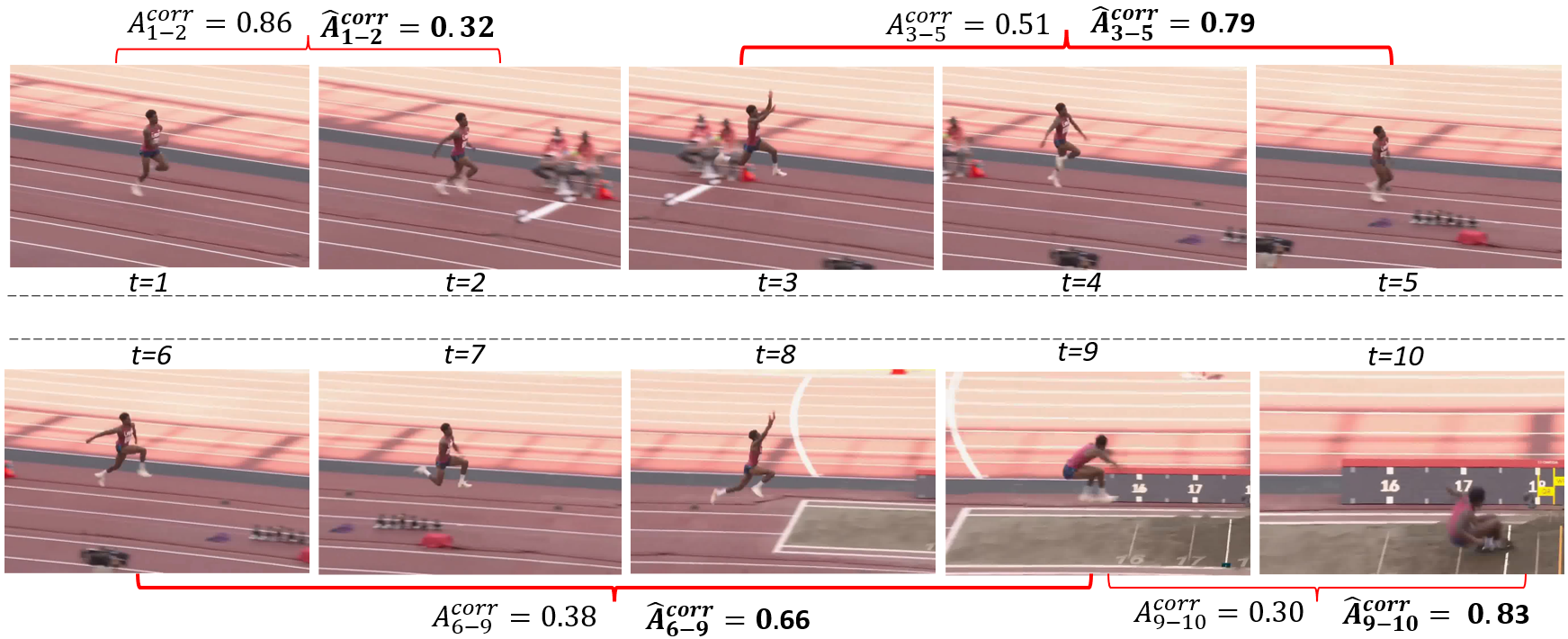}
	\caption{An example of action ``triple jump'' illustrates the comparison between the proposed sequence-based temporal attention correlation, $\hat{A}^{corr}$, and the standard temporal attention correlation, $A^{corr}$. Here, $A^{corr}_{i-j}$ and $\hat{A}^{corr}_{i-j}$ represents the temporal correlation between the frame $i$ and $j$.  The standard temporal attention tends to give higher values to similar frames, such as $t=1$ and $t=2$, and lower values to distinctive non-adjacent frames, such as $t=3$ and $t=5$ or $t=6$ and $t=9$. In contrast, the proposed temporal attention does not discriminate against frames based on their similarities. Hence, $\hat{A}^{corr}$  is more effective than $A^{corr}$ in representing the temporal dependencies between distinctive and non-adjacent frames that represent keyframes.
	\label{Fig:PTATT}} 	
\end{figure*}

Fig. \ref{Fig:PTATT} compares the proposed sequence-based temporal attention correlation and the standard one on an action sequence, ``triple jump'', with 10 sampled frames. Note that $A^{corr}$ and $\hat{A}^{corr}$ were explained in  (\ref{EQ:Temp1}) and  (\ref{EQ:Temp4}), respectively. Here, $A^{corr}_{i-j}$ and $\hat{A}^{corr}_{i-j}$ are the temporal correlation between the frame $i\in \tau$ and $j \in \tau$,
Investigation of Fig. \ref{Fig:PTATT} more closely indicates that the standard temporal attention model leans toward producing greater values to similar frames, such as $t=1$ and $t=2$. On the other hand,  The standard temporal attention sets lower values to distinctive and non-adjacent frames, such as $t=3$ and $t=5$ or $t=6$ and $t=9$. By contrast, the proposed sequence-based temporal attention does not discriminate among frames based on similarity. In other words, $\hat{A}^{corr}$  can more effectively provide the temporal relationship between distinctive and non-adjacent frames so-called keyframes.

\subsubsection{Enhancement of the current frameworks}

We integrated the proposed selective multi-feature attention (MFA) into several state-of-the-art spatiotemporal action detection frameworks based on the transformer network. This includes the Tublet attention in TubeR \cite{zhao2022CVPR}, actor-context attention in ACARNet \cite{pan2021CVPR}, and interaction attention in AIA \cite{tang2020ECCV}. The results for the AVA 2.1 dataset are shown in Table \ref{Tab:sota}. The outcomes indicated that all the methods improved when the new MFA was used, with the highest enhancement of 1.2\%.

\begin{table}[h!tbp]
	\centering
	\caption{The results of enhancing the state-of-the-art spatiotemporal action detection frameworks using the proposed selective multi-feature attention (MFA).} \label{Tab:sota}
\begin{tabular*}{\columnwidth}{@{\extracolsep{\fill}} ccc}
\hline
Method & mAP@0.5\% (baseline) & \textbf{mAP@0.5\% (MFA)}  \\ \hline
AIA \cite{tang2020ECCV} &  31.2 & \textbf{32.4}\\
ACARNet \cite{pan2021CVPR} & 30.0 & \textbf{30.1}\\
TubR \cite{zhao2022CVPR} & 32.0 & \textbf{32.5} \\
\hline
\end{tabular*}
\end{table}

\subsubsection{Error Analysis and Failure Cases}
Fig. \ref{Fig:vis1} (and Fig. 1 and Fig. 2 in the Supplementary Material) show some examples of success and failure cases from three sequences selected from the validation set of the AVA dataset. The true positives (TP), false positives (FP), and false negatives (FN) are the correct, incorrect, and missed action class predictions. 
Some \textbf{limitations} of the current pipeline that contributed to failure cases are (1) \emph{2D restriction of videos} is a cause of error. For example, the action ``walking along'' was incorrectly detected (FP) as  ``talk to'' or ``listen to'' due to the lack of 3D perception of face orientations. (2) \emph{Occlusion} between persons and objects can distract the network from detecting the correct relations between action semantics. For example, a background object occluded by persons led to the incorrect conclusion (FP) that the persons ``give''  objects to others. On the other hand, occlusion can also cause an object to be hidden, preventing it from detecting (FN) the action class label ``carry/hold'' an object. (3) \emph{The similarity between different classes} also has contributed to failure cases. For instance, the action ``sit'' could be confused with ``bend/bow''.

\begin{figure*}[h!tbp]
	\centering
	\includegraphics[height=0.48\textwidth]{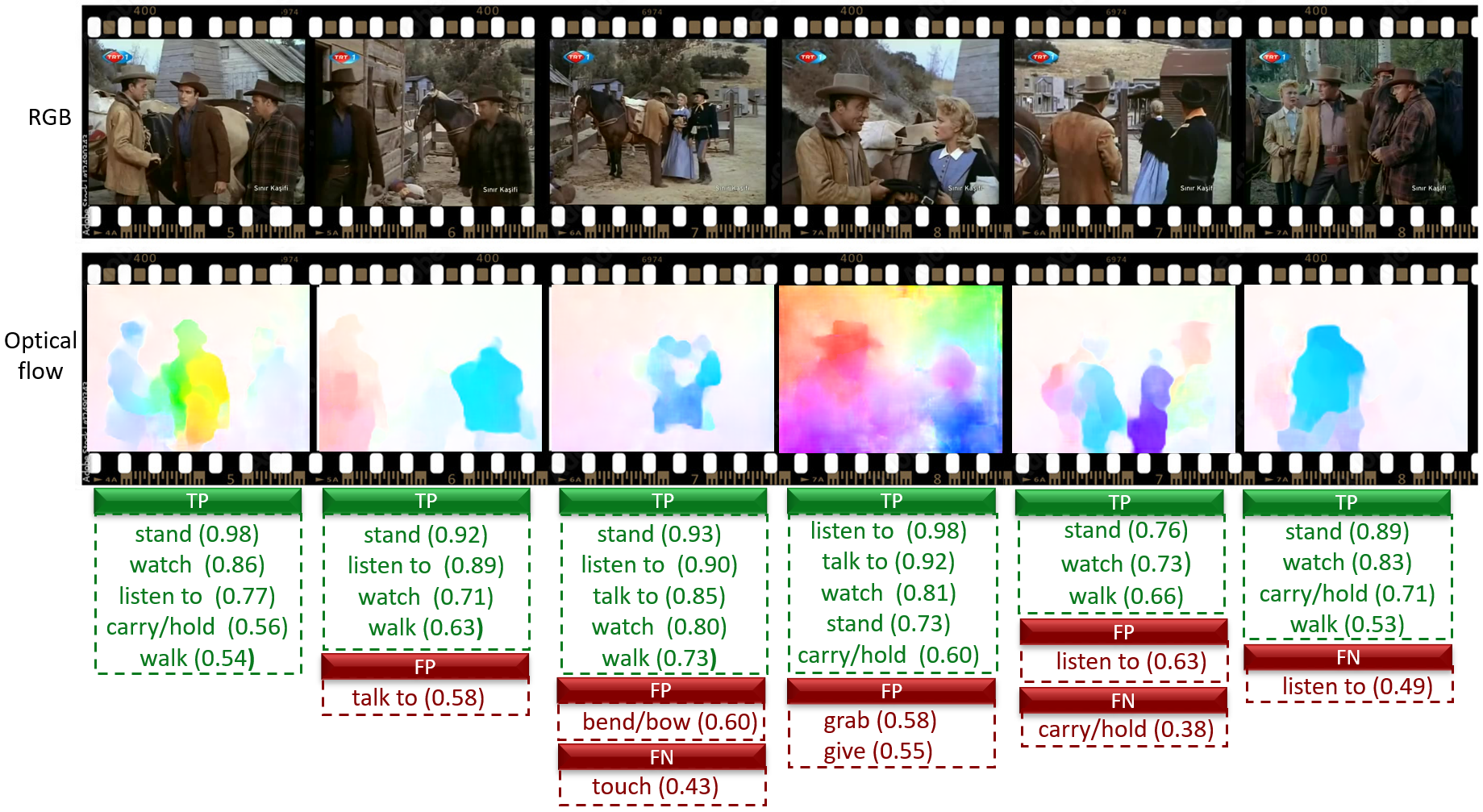} \\
 video id: ``7T5G0CmwTPo'' \\
	\caption{Examples of success and failure cases on a sequence of the AVA validation set (video id: ``7T5G0CmwTPo''). The results show the true positive (TP), false positive (FP), and false negative (FN) predicted classes with their confidence scores.  
	\label{Fig:vis1}} 	
\end{figure*}

\subsubsection{Computational Efficiency Analysis}

Table \ref{Tab:time_anal} shows the efficiency analysis of different pipeline modules, based on running our algorithm on 30 frames sampled from 5 seconds of an action video. Furthermore, Table \ref{Tab:time_anal2} indicates the efficiency analysis of different combinations of multi-feature attention types.

\begin{table}[h!tbp]
	\centering
	\caption{\textcolor{black}{Efficiency analysis of different pipeline modules.} } \label{Tab:time_anal} 
\begin{tabular}{cc}
\hline
Module & time (ms) \\ \hline
Action semantic detection &  401 \\ \hline
Optical Flow extraction &   233 \\ \hline
Motion Memory Module  &  106 \\ \hline
Multi-feature selective semantic attention & 160 \\ \hline
Multi-feature regular semantic attention & 235 \\ \hline
Temporal Attention & 41  \\ \hline
Classification and Regression & 35  \\ \hline
Total pre-processing & 634  \\ \hline
Total network modules & 342 \\ \hline
Total runtime & 976  \\
 \hline
\end{tabular}
\end{table}

\begin{table}[h!tbp]
	\centering
	\caption{\textcolor{black}{Efficiency analysis of different multi-feature attention (MFA) modules and their combinations.} } \label{Tab:time_anal2} 
\begin{tabular}{cc}
\hline
MFA Module & time (ms) \\ \hline
$A^{GG}$ &  101 \\ \hline
$A^{MM}$ &   108  \\ \hline
$A^{GM} + A^{MG}$ &  143 \\ \hline
$A^{GG}$ + $A^{MM}$ +  $A^{GM} + A^{MG}$ & 160\\  \hline

 \hline
\end{tabular}
\end{table}


\section{Conclusions and Future Work}
This paper presents a novel spatiotemporal transformer network to model action semantics and their interactions effectively. In terms of contributed components, we introduce, first, a motion-aware 2D positional encoding that, in contrast to the standard one, can handle spatiotemporal variations in videos, especially geared toward action semantics. Second, a multi-feature attention model captures complicated multi-feature interactions between action semantics based on their spatial and motion properties. This model is accompanied by a special-purpose selective attention mechanism, which is more effective and efficient than the standard self-attention mechanism. Third, the sequence-based temporal attention model effectively captures heterogeneous temporal dependencies between action frames. In contrast to the traditional temporal attention mechanism that focuses on the relationship between individual action frames, the sequence-based temporal attention model prioritizes the temporal dependencies between frames based on their contribution to the action sequence. 
The proposed pipeline outperformed the state-of-the-art methods on four spatiotemporal action benchmarks: AVA 2.2, AVA 2.1, UCF101-24, and EPIC-Kitchens.

The current pipeline detects action semantics with a pre-trained network, which is not end-to-end. A seamless learning process within the transformer network that encodes action semantics directly may be more desirable. The same criticism could also be levied toward our current method of encoding optical flow fields. As possible future work, we suggest extracting or simulating the optical flow fields end-to-end within the deep network.

\section{Supplementary Material}

\subsection{Extended Algorithm Summary}
Table \ref{Tab:alg_sum_ext} indicates the extended algorithm summary.

\begin{table*}[h!tbp]
\footnotesize
	\centering
	\caption{\textcolor{black}{Pipeline algorithm summary in a hierarchical order indicating each phase (in bold), the summary of each phase, inputs and outputs of each phase, and the corresponding manuscript section}.} \label{Tab:alg_sum_ext} 
\begin{tabular}{cccc}
\hline
Phase/Summary & Inputs & Outputs & Section \\ \hline
\textbf{semantic detection} & $I^{RGB}$ & $S^G$ &3.2 \\ 
extracts action semantics (people and objects) & RGB frames & spatial action semantics & \\ \hline 
\textbf{optical flow extraction}  & $I^{RGB}$ &  $S^M$, $S^V$ & 3.2 \\ 
extracts optical flow fields for action semantics & RGB frames & motion semantics images, and vectors & \\ \hline
\textbf{motion enhancement and segmentation} & $B^G$, $I^V$ & $S^M$, $S^V$ & 3.2.1 \\
improves motion features (w.r.t camera) &  bounding boxes, motion vectors & enhanced motion semantics, and vectors \\ \hline
\textbf{feature embedding} & $S^G$, $S^M$ & $X^G$, $X^M$ & 3.3\\ 
extract spatial and motion features using Conv & spatial, and motion semantics & spatial, and motion feature embeddings & \\ \hline
\textbf{motion memory module} & $S^V$ & $\Delta P^X$, $\Delta P^Y$ & 3.3.1 \\ 
provides motion information to the transformer & semantic motion vectors & semantic motion offsets & \\ \hline
\textbf{motion-aware 2D Positional Encoding} & $\Delta P^X$, $\Delta P^Y$ & $P^X_A$, $P^X_Y$ & 3.3.1  \\
makes the transformer aware of movements & semantic motion offsets & 2D motion-aware positional encoding & \\ \hline
\textbf{multi-feature semantic attention} & $S^G$, $S^M$ & $A^H$ & 3.3.2  \\
computes the correlations between multi-features & spatial, and motion features& multi-head semantic attention & \\ \hline
\textbf{multi-feature fusion} & $A^H$ & $X_s$ & 3.3.3 \\ 
combines and flows the features in different layers & multi-head semantic attention, & temporal semantic features & \\ \hline
\textbf{sequence-based temporal attention} & $X_s$ & $\hat{A}$ & 3.3.4 \\ 
computes heterogeneous temporal relations & temporal semantic features & temporal attention & \\ \hline
\textbf{classification and regression} & $\hat{A}$ & $\hat{Y}$, $t_s$, $t_e$ & 3.3.5 \\
classifies actions and start and end frames & temporal attention  & class predictions, start and end frames & \\
\hline
\end{tabular}
\end{table*}

\subsection{Semantic Motion Enhancement and Segmentation}
\emph{Motion modeling:}  In this step, the background motion $S^{V}_B(\lambda) = \{\lambda_n, n \in 0,..., H\}$ is modeled, where $\lambda=(u, v)$, and $H$ is the size of $S^{V}_B(\lambda)$. We assume that $S^{V}_B(\lambda) = S^{V}(\lambda) - S^{V}_Z(\lambda)$, where $S^{V}_Z(\lambda) \in S^{V}$ are the semantic motion vectors for persons that are segmented from $S^{V}(\lambda)$ using the corresponding bounding boxes for persons, $B^Z$.
The Gaussian mixture models (GMMs) with $K$ distributions is used to model the background motion as $P(\lambda)  = \sum_{k=1}^{K} \pi_k N(\lambda | \mu_k, \Sigma_k)$, where $N(\lambda|\mu_k, \Sigma_k)$ is a Gaussian density with the mean of $\mu_k$ and the covariance of $\Sigma_m$, and mixing coefficient of $\pi_k$. $S^{V}_B(\lambda)$ is modeled by optimizing the GMMs parameters by way of the maximum likelihood estimation for the GMMs. 

\emph{Motion restoration:}
In this step, the semantic motions vectors $S^{V} = \{S^{V}_Z, S^{V}_O\} \in S^{V}$, which includes the vectors for persons, $S^{V}_Z$, and objects, $S^{V}_O$ are restored. Each motion vector is restored as: $\lambda_n \in S^{V} = \lambda_n - \mu_m$, where $\lambda_n$ belongs to one of the $k \in K$ GMMs distributions.  The final enhanced segmented semantic motion vectors $S^{M}_A = \{S^{V}, S^{M}$\} are used as reliable motion representations of action semantics in the spatiotemporal transformer network, where  $S^{M}$ is the corresponding segmented semantic motion images that are converted from vectors. 

\begin{figure*}[!htbp]
\renewcommand{\tabcolsep}{0.5pt}
	\centering 
		\begin{tabular}{cccc}
   \includegraphics[height=0.15\textheight]{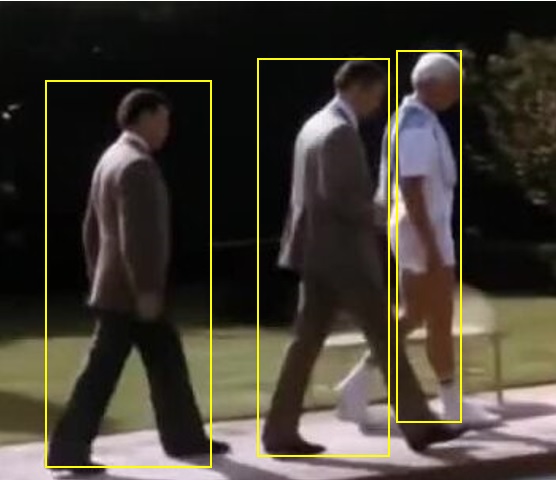} &
  \includegraphics[height=0.15\textheight]{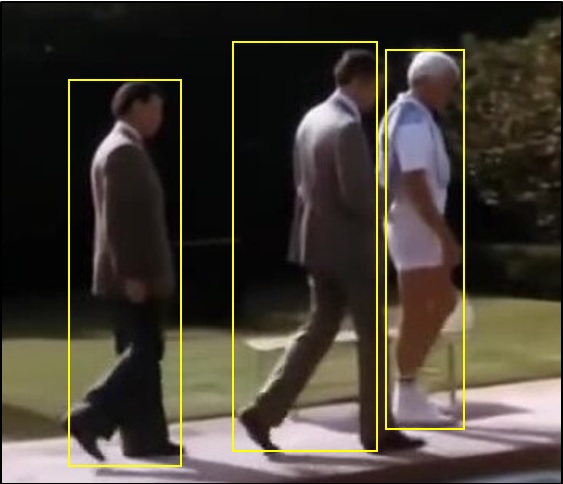} &
  \includegraphics[height=0.15\textheight]{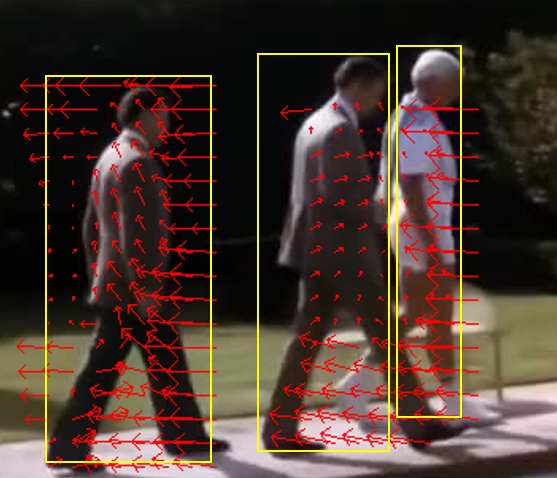} &
  \includegraphics[height=0.15\textheight]{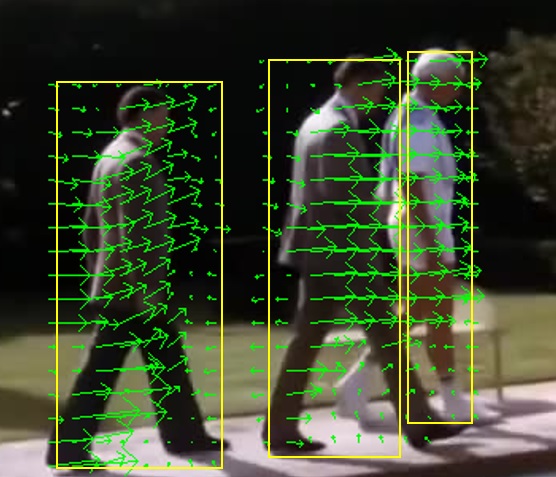} \\
      time: $t$ & time: $t + \omega$ &
  distorted segmented & our enhanced segmented \\ 
  & & semantic motion vectors & semantic motion vectors \\
  	\end{tabular}
	\caption{An example of the semantic motion enhancement and segmentation algorithm tested on an action sample from the AVA dataset. For two consecutive sampled frames of $t$ and $t + \omega$, the distorted semantic motion vectors (red arrows) and the enhanced semantic motion vectors (green arrows) are shown. In this example, the segmented semantics (persons) walk to the right. At the same time, the camera also moves to the right (marginally faster), causing incorrect extraction of semantic motion vectors. The proposed motion enhancement and segmentation algorithm fixes this issue.  ~\label{Fig:exam_act4}}
\end{figure*}

Some examples are shown in Fig. \ref{Fig:exam_act4} (raw images are selected from the AVA dataset). In this example, the persons walk to their right as the camera moves in the same direction slightly faster. Such spatiotemporal inconsistency leads to incorrect estimation of motion vectors showing the wrong motion direction to the left (red motion vectors). As a result, the enhanced segmented semantic motion vectors are illustrated in Fig. \ref{Fig:exam_act4}, in which the distorted and resorted motion vectors are shown in red and green, respectively.

\subsection{Proof for Efficiency of Selective Attention}
The following show the matrix multiplications between query ($Q$), key ($K$) and value ($V$) of our selective semantic attention ($A^S$) that are illustrated with a size operation ($d$):

For the selective semantic attention, $A^S$, the dimensionality and computational cost are given by:

\begin{align}
\begin{split}
    d(A^S) &= d(Q) \times d(K)^T \times d(V) \\
    &= \mathbb{R}^{N \times d_h}  \times (\mathbb{R}^{(N+N') \times d_h})^T \times \mathbb{R}^{(N+N') \times d_h} \\
    &= \mathbb{R}^{N \times d_h}  \times \mathbb{R}^{d_h \times (N+N') } \times \mathbb{R}^{(N+N') \times d_h} \\
    &= \mathbb{R}^{N \times d_h}.
\end{split}
\end{align}
This makes the computational cost of our selective semantic attention $O(N \cdot (N + N') \cdot d_h)$.

For regular semantic self-attention, $A^R$, the dimensions are:
\begin{align}
\begin{split}
    d(A^R) &= d(Q) \times d(K)^T \times d(V) \\
    &= \mathbb{R}^{(N+N') \times d_h}  \times (\mathbb{R}^{(N+N') \times d_h})^T \times \mathbb{R}^{(N+N') \times d_h} \\
    &= \mathbb{R}^{(N+N') \times d_h}  \times \mathbb{R}^{d_h \times (N+N') } \times \mathbb{R}^{(N+N') \times d_h} \\
    &= \mathbb{R}^{(N+N') \times d_h}.
\end{split}
\end{align}

So, the computational cost of the regular semantic self-attention is $O((N + N')^2 \cdot d_h)$.

\subsection{Sequence-based temporal Attention}

\subsubsection{Conceptual Proof}
Let's think of the action frames as employees and the action as a company. Our approach is to value the professional relationships (temporal dependencies) between employees (frames) that benefit the company (action), such as effective teamwork, rather than personal relationships, such as liking the same foods. We also can prove the above conclusion by way of \emph{maximum a posteriori estimation}. Specifically, given a class $c$ and observation $x$ that could be a frame $f^s_t \in X_s$ or an action sequence $X_s$, the goal of the action detection network is to maximize the posterior of $p(c|x) = \frac{p(x|c)}{p(x)} \cdot p(c)$. We can say that the likelihood  $p(X_s|c) > p(f^s_t|c)$ since a sequence better represents an action than a frame. On the other hand, $p(X_s) < p(f^s_t)$ since having a unique combination of frames is less common than individual frames. So, $p(c|X_s) > p(f_t^s)$.

\subsubsection{Mathematical Proof}
The sequence-based temporal attention is based on the intuition of the Mahalanobis distance, which determines the distance between two vectors, $q$ and $k$, with respect to the data distribution $X_s$ as $D(q, k;X_s) = \sqrt{(q-k)S^{-1}(q-k)^T}$. Assuming the correlations between two vectors, $corr(q, k)$ can be defined as $q.k^T = 1 - D^2(q^2, k^2)/2N$, by replacing it in  (21), we can justify (23). We, however, instead of the vector product used in computing the Mahalanobis distance, perform a matrix multiplication between vectors to calculate the frame-by-frame sequence-based attention correlation matrix, $\hat{A}^{corr}$.

\subsection{Visualization of Predicted Results}
\begin{figure*}[h!tbp]
	\centering
	\includegraphics[height=0.49\textwidth]{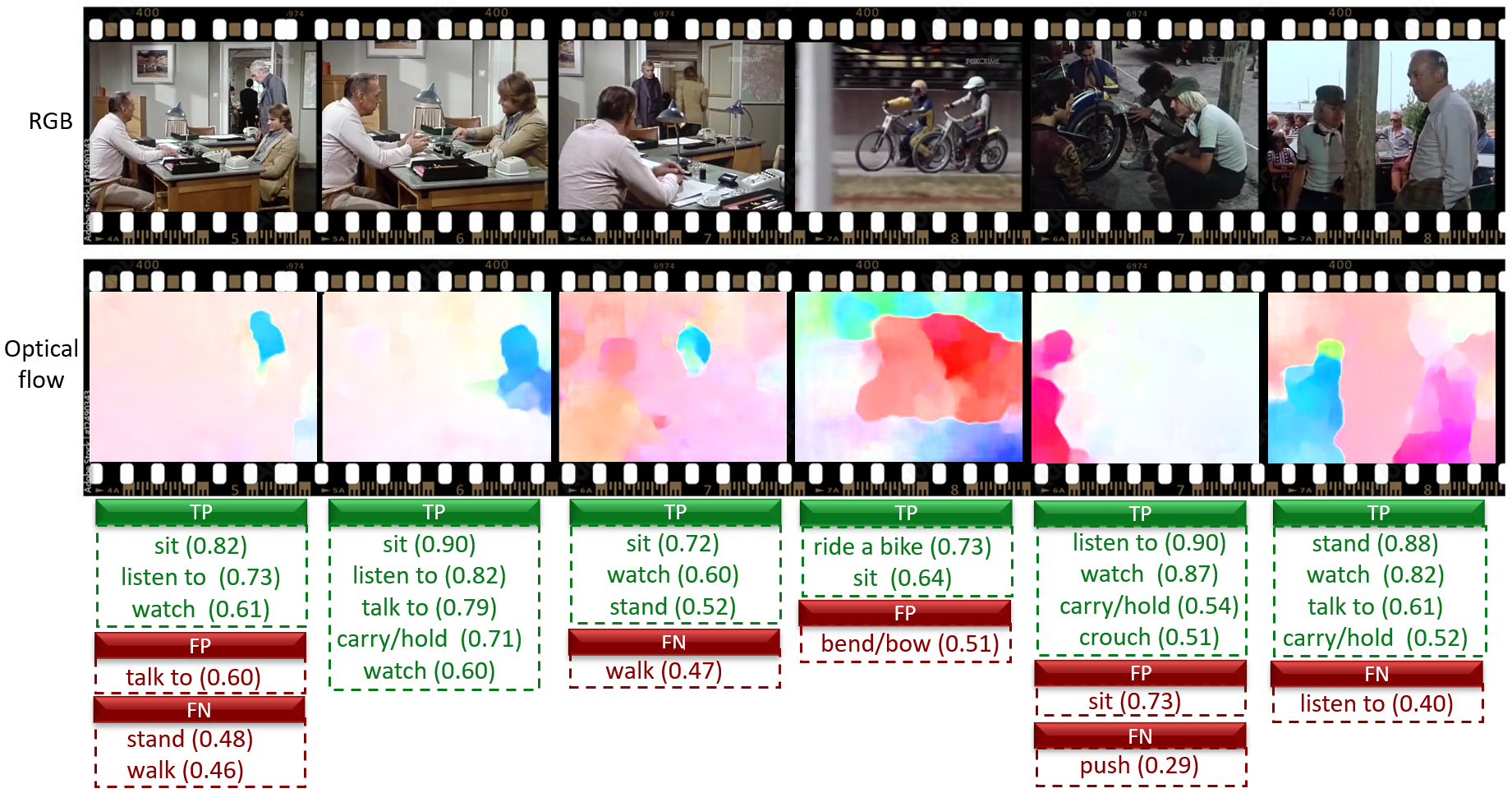} \\
 video id: ``CZ2NP8UsPuE'' \\
	\caption{Some examples of success and failure scenarios from an action sequence of the validation set of the AVA dataset (video id: ``CZ2NP8UsPuE''). The results indicate the true positive (TP), false positive (FP), and false negative (FN) predicted classes with their confidence scores.
	\label{Fig:vis6}} 	
\end{figure*}

\begin{figure*}[h!tbp]
	\centering
	\includegraphics[height=0.49\textwidth]{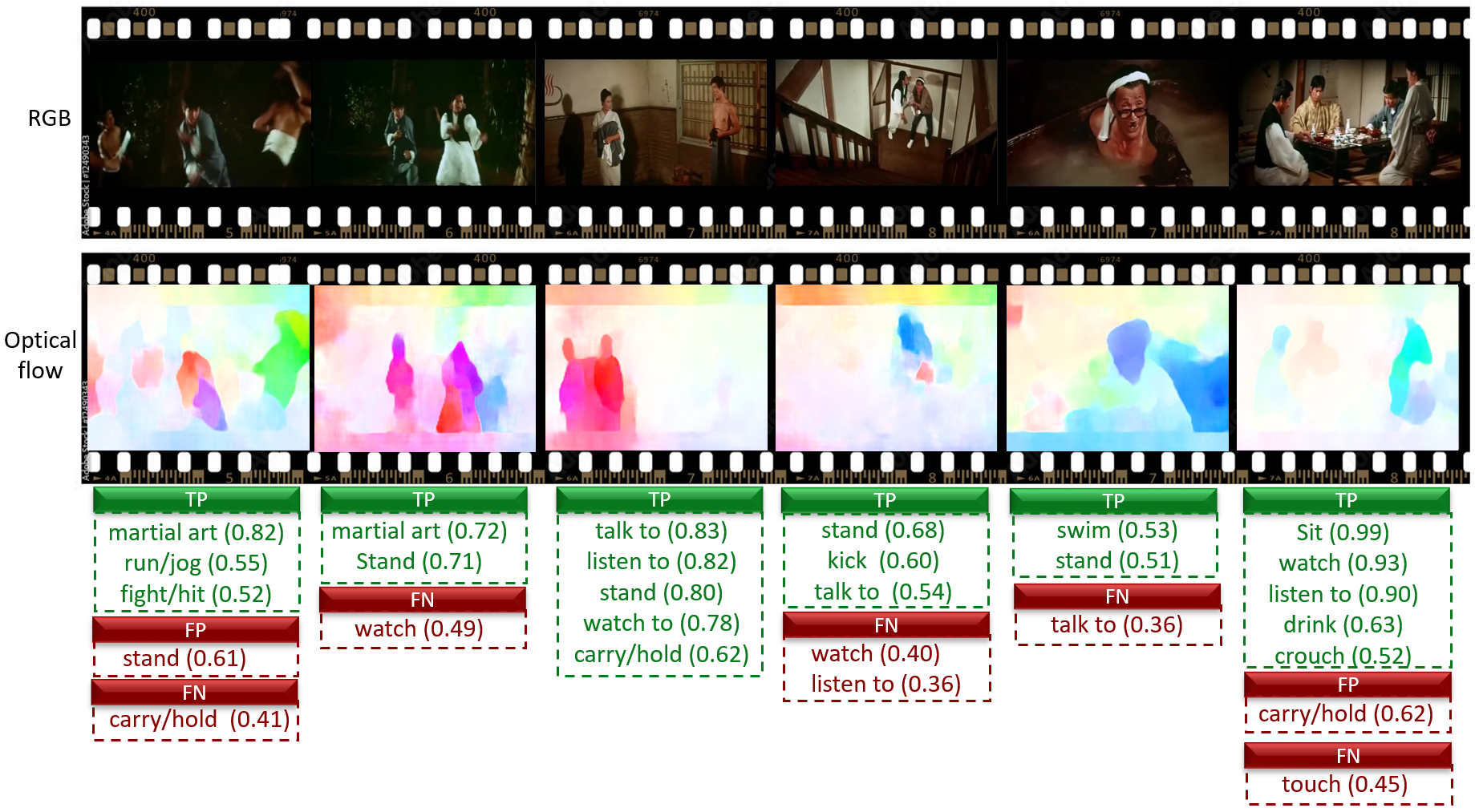} \\
 video id: ``Hscyg0vLKc8'' \\
	\caption{An example sequence of the AVA validation set showing the success and failure cases (video id: ``Hscyg0vLKc8''). The results are true positive (TP), false positive (FP), and false negative (FN) predicted classes with their confidence scores.
	\label{Fig:vis7}} 	
\end{figure*}

Fig. \ref{Fig:vis6} and Fig. \ref{Fig:vis7} show some examples of success and failure cases from three sequences selected from the validation set of the AVA dataset. The true positives (TP), false positives (FP), and false negatives (FN) are the correct, incorrect, and missed action class predictions.

\bibliographystyle{unsrt}
\bibliography{bibfile}

\begin{thebibliography}{10}

\bibitem{oikonomopoulos2010TIP}
Antonios Oikonomopoulos, Ioannis Patras, and Maja Pantic.
\newblock Spatiotemporal localization and categorization of human actions in unsegmented image sequences.
\newblock {\em IEEE transactions on Image Processing}, 20(4):1126--1140, 2010.

\bibitem{tran2012NIPS}
Du~Tran and Junsong Yuan.
\newblock Max-margin structured output regression for spatio-temporal action localization.
\newblock {\em Advances in neural information processing systems}, 25, 2012.

\bibitem{weinzaepfel2015ICCV}
Philippe Weinzaepfel, Zaid Harchaoui, and Cordelia Schmid.
\newblock Learning to track for spatio-temporal action localization.
\newblock In {\em Proceedings of the IEEE international conference on computer vision}, pages 3164--3172, 2015.

\bibitem{girdhar2019CVPR}
Rohit Girdhar, Joao Carreira, Carl Doersch, and Andrew Zisserman.
\newblock Video action transformer network.
\newblock In {\em Proceedings of the IEEE/CVF conference on computer vision and pattern recognition}, pages 244--253, 2019.

\bibitem{li2020ATIP}
Bin Li, Jian Tian, Zhongfei Zhang, Hailin Feng, and Xi~Li.
\newblock Multitask non-autoregressive model for human motion prediction.
\newblock {\em IEEE Transactions on Image Processing}, 30:2562--2574, 2020.

\bibitem{vaswani2017NIPS}
Ashish Vaswani, Noam Shazeer, Niki Parmar, Jakob Uszkoreit, Llion Jones, Aidan~N Gomez, {\L}ukasz Kaiser, and Illia Polosukhin.
\newblock Attention is all you need.
\newblock {\em Advances in neural information processing systems}, 30, 2017.

\bibitem{carion2020ECCV}
Nicolas Carion, Francisco Massa, Gabriel Synnaeve, Nicolas Usunier, Alexander Kirillov, and Sergey Zagoruyko.
\newblock End-to-end object detection with transformers.
\newblock In {\em European conference on computer vision}, pages 213--229. Springer, 2020.

\bibitem{Dosovitskiy2020ICLR}
Alexey Dosovitskiy, Lucas Beyer, Alexander Kolesnikov, Dirk Weissenborn, Xiaohua Zhai, Thomas Unterthiner, Mostafa Dehghani, Matthias Minderer, Georg Heigold, Sylvain Gelly, et~al.
\newblock An image is worth 16x16 words: Transformers for image recognition at scale.
\newblock {\em arXiv preprint arXiv:2010.11929}, 2020.

\bibitem{korban2020ECCV}
Matthew Korban and Xin Li.
\newblock Ddgcn: A dynamic directed graph convolutional network for action recognition.
\newblock In {\em Computer Vision--ECCV 2020: 16th European Conference, Glasgow, UK, August 23--28, 2020, Proceedings, Part XX 16}, pages 761--776. Springer, 2020.

\bibitem{korban2023PR2}
Matthew Korban and Xin Li.
\newblock Semantics-enhanced early action detection using dynamic dilated convolution.
\newblock {\em Pattern Recognition}, 140:109595, 2023.

\bibitem{guo2013TIP}
Kai Guo, Prakash Ishwar, and Janusz Konrad.
\newblock Action recognition from video using feature covariance matrices.
\newblock {\em IEEE Transactions on Image Processing}, 22(6):2479--2494, 2013.

\bibitem{khan2014TIP}
Fahad~Shahbaz Khan, Joost Van De~Weijer, Rao~Muhammad Anwer, Michael Felsberg, and Carlo Gatta.
\newblock Semantic pyramids for gender and action recognition.
\newblock {\em IEEE transactions on image processing}, 23(8):3633--3645, 2014.

\bibitem{kalogeiton2017ICCV}
Vicky Kalogeiton, Philippe Weinzaepfel, Vittorio Ferrari, and Cordelia Schmid.
\newblock Action tubelet detector for spatio-temporal action localization.
\newblock In {\em Proceedings of the IEEE International Conference on Computer Vision}, pages 4405--4413, 2017.

\bibitem{wang2020PAMI}
Xiaohan Wang, Linchao Zhu, Yu~Wu, and Yi~Yang.
\newblock Symbiotic attention for egocentric action recognition with object-centric alignment.
\newblock {\em IEEE transactions on pattern analysis and machine intelligence}, 2020.

\bibitem{pan2021CVPR}
Junting Pan, Siyu Chen, Mike~Zheng Shou, Yu~Liu, Jing Shao, and Hongsheng Li.
\newblock Actor-context-actor relation network for spatio-temporal action localization.
\newblock In {\em Proceedings of the IEEE/CVF Conference on Computer Vision and Pattern Recognition}, pages 464--474, 2021.

\bibitem{simonyan2014NIPS}
Karen Simonyan and Andrew Zisserman.
\newblock Two-stream convolutional networks for action recognition in videos.
\newblock {\em Advances in neural information processing systems}, 27, 2014.

\bibitem{feichtenhofer2016CVPR}
Christoph Feichtenhofer, Axel Pinz, and Andrew Zisserman.
\newblock Convolutional two-stream network fusion for video action recognition.
\newblock In {\em Proceedings of the IEEE conference on computer vision and pattern recognition}, pages 1933--1941, 2016.

\bibitem{wang2017MM}
Xuanhan Wang, Lianli Gao, Peng Wang, Xiaoshuai Sun, and Xianglong Liu.
\newblock Two-stream 3-d convnet fusion for action recognition in videos with arbitrary size and length.
\newblock {\em IEEE Transactions on Multimedia}, 20(3):634--644, 2017.

\bibitem{su2020PAMI}
Rui Su, Dong Xu, Luping Zhou, and Wanli Ouyang.
\newblock Progressive cross-stream cooperation in spatial and temporal domain for action localization.
\newblock {\em IEEE Transactions on Pattern Analysis and Machine Intelligence}, 43(12):4477--4490, 2020.

\bibitem{Hu2022PAMI}
Weiming Hu, Haowei Liu, Yang Du, Chunfeng Yuan, Bing Li, and Stephen Maybank.
\newblock Interaction-aware spatio-temporal pyramid attention networks for action classification.
\newblock {\em IEEE Transactions on Pattern Analysis and Machine Intelligence}, 44(10):7010--7028, 2022.

\bibitem{singh2017ICCV}
Gurkirt Singh, Suman Saha, Michael Sapienza, Philip~HS Torr, and Fabio Cuzzolin.
\newblock Online real-time multiple spatiotemporal action localisation and prediction.
\newblock In {\em Proceedings of the IEEE International Conference on Computer Vision}, pages 3637--3646, 2017.

\bibitem{yang2019CVPR}
Xitong Yang, Xiaodong Yang, Ming-Yu Liu, Fanyi Xiao, Larry~S Davis, and Jan Kautz.
\newblock Step: Spatio-temporal progressive learning for video action detection.
\newblock In {\em Proceedings of the IEEE/CVF Conference on Computer Vision and Pattern Recognition}, pages 264--272, 2019.

\bibitem{singh2023WCACV}
Gurkirt Singh, Vasileios Choutas, Suman Saha, Fisher Yu, and Luc Van~Gool.
\newblock Spatio-temporal action detection under large motion.
\newblock In {\em Proceedings of the IEEE/CVF Winter Conference on Applications of Computer Vision}, pages 6009--6018, 2023.

\bibitem{kong2021ICRA}
Lingtong Kong, Chunhua Shen, and Jie Yang.
\newblock Fastflownet: A lightweight network for fast optical flow estimation.
\newblock In {\em 2021 IEEE International Conference on Robotics and Automation (ICRA)}, pages 10310--10316. IEEE, 2021.

\bibitem{korban2023PR}
Matthew Korban, Peter Youngs, and Scott~T Acton.
\newblock A multi-modal transformer network for action detection.
\newblock {\em Pattern Recognition}, 142:109713, 2023.

\bibitem{Liu2022TIP}
Xiaolong Liu, Qimeng Wang, Yao Hu, Xu~Tang, Shiwei Zhang, Song Bai, and Xiang Bai.
\newblock End-to-end temporal action detection with transformer.
\newblock {\em IEEE Transactions on Image Processing}, 31:5427--5441, 2022.

\bibitem{khan2022ACS}
Salman Khan, Muzammal Naseer, Munawar Hayat, Syed~Waqas Zamir, Fahad~Shahbaz Khan, and Mubarak Shah.
\newblock Transformers in vision: A survey.
\newblock {\em ACM computing surveys (CSUR)}, 54(10s):1--41, 2022.

\bibitem{han2021NIPS}
Kai Han, An~Xiao, Enhua Wu, Jianyuan Guo, Chunjing Xu, and Yunhe Wang.
\newblock Transformer in transformer.
\newblock {\em Advances in Neural Information Processing Systems}, 34:15908--15919, 2021.

\bibitem{korban2023PR3}
Matthew Korban, Peter Youngs, and Scott~T Acton.
\newblock Taa-gcn: A temporally aware adaptive graph convolutional network for age estimation.
\newblock {\em Pattern Recognition}, 134:109066, 2023.

\bibitem{ballakur2020ICCCS}
Amulya~Arun Ballakur and Arti Arya.
\newblock Empirical evaluation of gated recurrent neural network architectures in aviation delay prediction.
\newblock In {\em 2020 5th International Conference on Computing, Communication and Security (ICCCS)}, pages 1--7. IEEE, 2020.

\bibitem{soomro2012ucf101}
Khurram Soomro, Amir~Roshan Zamir, and Mubarak Shah.
\newblock Ucf101: A dataset of 101 human actions classes from videos in the wild.
\newblock {\em arXiv preprint arXiv:1212.0402}, 2012.

\bibitem{kalyan2021JBI}
Katikapalli~Subramanyam Kalyan, Ajit Rajasekharan, and Sivanesan Sangeetha.
\newblock Ammu: a survey of transformer-based biomedical pretrained language models.
\newblock {\em Journal of biomedical informatics}, page 103982, 2021.

\bibitem{gabbur2021NIPS}
Prasad Gabbur, Manjot Bilkhu, and Javier Movellan.
\newblock Probabilistic attention for interactive segmentation.
\newblock {\em Advances in Neural Information Processing Systems}, 34:4448--4460, 2021.

\bibitem{nguyen2022ICML}
Tam~Minh Nguyen, Tan~Minh Nguyen, Dung~DD Le, Duy~Khuong Nguyen, Viet-Anh Tran, Richard Baraniuk, Nhat Ho, and Stanley Osher.
\newblock Improving transformers with probabilistic attention keys.
\newblock In {\em International Conference on Machine Learning}, pages 16595--16621. PMLR, 2022.

\bibitem{abd2003ICS}
Wael Abd-Almageed, Aly El-Osery, and Christopher~E Smith.
\newblock Non-parametric expectation maximization: a learning automata approach.
\newblock In {\em SMC'03 Conference Proceedings. 2003 IEEE International Conference on Systems, Man and Cybernetics. Conference Theme-System Security and Assurance (Cat. No. 03CH37483)}, volume~3, pages 2996--3001. IEEE, 2003.

\bibitem{mayer2016CVPR}
Nikolaus Mayer, Eddy Ilg, Philip Hausser, Philipp Fischer, Daniel Cremers, Alexey Dosovitskiy, and Thomas Brox.
\newblock A large dataset to train convolutional networks for disparity, optical flow, and scene flow estimation.
\newblock In {\em Proceedings of the IEEE conference on computer vision and pattern recognition}, pages 4040--4048, 2016.

\bibitem{damen2018ECCV}
Dima Damen, Hazel Doughty, Giovanni~Maria Farinella, Sanja Fidler, Antonino Furnari, Evangelos Kazakos, Davide Moltisanti, Jonathan Munro, Toby Perrett, Will Price, et~al.
\newblock Scaling egocentric vision: The epic-kitchens dataset.
\newblock In {\em Proceedings of the European Conference on Computer Vision (ECCV)}, pages 720--736, 2018.

\bibitem{lin2017ICCV}
Tsung-Yi Lin, Priya Goyal, Ross Girshick, Kaiming He, and Piotr Doll{\'a}r.
\newblock Focal loss for dense object detection.
\newblock In {\em Proceedings of the IEEE international conference on computer vision}, pages 2980--2988, 2017.

\bibitem{gu2018CVPR}
Chunhui Gu, Chen Sun, David~A Ross, Carl Vondrick, Caroline Pantofaru, Yeqing Li, Sudheendra Vijayanarasimhan, George Toderici, Susanna Ricco, Rahul Sukthankar, et~al.
\newblock Ava: A video dataset of spatio-temporally localized atomic visual actions.
\newblock In {\em Proceedings of the IEEE Conference on Computer Vision and Pattern Recognition}, pages 6047--6056, 2018.

\bibitem{tang2020ECCV}
Jiajun Tang, Jin Xia, Xinzhi Mu, Bo~Pang, and Cewu Lu.
\newblock Asynchronous interaction aggregation for action detection.
\newblock In {\em Computer Vision--ECCV 2020: 16th European Conference, Glasgow, UK, August 23--28, 2020, Proceedings, Part XV 16}, pages 71--87. Springer, 2020.

\bibitem{wu2019CVPR}
Chao-Yuan Wu, Christoph Feichtenhofer, Haoqi Fan, Kaiming He, Philipp Krahenbuhl, and Ross Girshick.
\newblock Long-term feature banks for detailed video understanding.
\newblock In {\em Proceedings of the IEEE/CVF Conference on Computer Vision and Pattern Recognition}, pages 284--293, 2019.

\bibitem{baradel2018ECCV}
Fabien Baradel, Natalia Neverova, Christian Wolf, Julien Mille, and Greg Mori.
\newblock Object level visual reasoning in videos.
\newblock In {\em Proceedings of the European Conference on Computer Vision (ECCV)}, pages 105--121, 2018.

\bibitem{girdhar2022omnivore_cvpr}
Rohit Girdhar, Mannat Singh, Nikhila Ravi, Laurens van~der Maaten, Armand Joulin, and Ishan Misra.
\newblock Omnivore: A single model for many visual modalities.
\newblock In {\em Proceedings of the IEEE/CVF Conference on Computer Vision and Pattern Recognition}, pages 16102--16112, 2022.

\bibitem{wu2022CVPR}
Chao-Yuan Wu, Yanghao Li, Karttikeya Mangalam, Haoqi Fan, Bo~Xiong, Jitendra Malik, and Christoph Feichtenhofer.
\newblock Memvit: Memory-augmented multiscale vision transformer for efficient long-term video recognition.
\newblock In {\em Proceedings of the IEEE/CVF Conference on Computer Vision and Pattern Recognition}, pages 13587--13597, 2022.

\bibitem{herzig2022CVPR}
Roei Herzig, Elad Ben-Avraham, Karttikeya Mangalam, Amir Bar, Gal Chechik, Anna Rohrbach, Trevor Darrell, and Amir Globerson.
\newblock Object-region video transformers.
\newblock In {\em Proceedings of the IEEE/CVF Conference on Computer Vision and Pattern Recognition}, pages 3148--3159, 2022.

\bibitem{tong2022NIPS}
Zhan Tong, Yibing Song, Jue Wang, and Limin Wang.
\newblock Videomae: Masked autoencoders are data-efficient learners for self-supervised video pre-training.
\newblock {\em Advances in neural information processing systems}, 35:10078--10093, 2022.

\bibitem{feichten2021CVPR}
Christoph Feichtenhofer, Haoqi Fan, Bo~Xiong, Ross Girshick, and Kaiming He.
\newblock A large-scale study on unsupervised spatiotemporal representation learning.
\newblock In {\em Proceedings of the IEEE/CVF Conference on Computer Vision and Pattern Recognition}, pages 3299--3309, 2021.

\bibitem{fan2021ICCV}
Haoqi Fan, Bo~Xiong, Karttikeya Mangalam, Yanghao Li, Zhicheng Yan, Jitendra Malik, and Christoph Feichtenhofer.
\newblock Multiscale vision transformers.
\newblock In {\em Proceedings of the IEEE/CVF International Conference on Computer Vision}, pages 6824--6835, 2021.

\bibitem{feichtenhofer2022CVPR}
Christoph Feichtenhofer.
\newblock X3d: Expanding architectures for efficient video recognition.
\newblock In {\em Proceedings of the IEEE/CVF conference on computer vision and pattern recognition}, pages 203--213, 2020.

\bibitem{chen2021ICCV}
Shoufa Chen, Peize Sun, Enze Xie, Chongjian Ge, Jiannan Wu, Lan Ma, Jiajun Shen, and Ping Luo.
\newblock Watch only once: An end-to-end video action detection framework.
\newblock In {\em Proceedings of the IEEE/CVF International Conference on Computer Vision}, pages 8178--8187, 2021.

\bibitem{wu2021CVPR}
Chao-Yuan Wu and Philipp Krahenbuhl.
\newblock Towards long-form video understanding.
\newblock In {\em Proceedings of the IEEE/CVF Conference on Computer Vision and Pattern Recognition}, pages 1884--1894, 2021.

\bibitem{zhao2022CVPR}
Jiaojiao Zhao, Yanyi Zhang, Xinyu Li, Hao Chen, Bing Shuai, Mingze Xu, Chunhui Liu, Kaustav Kundu, Yuanjun Xiong, Davide Modolo, et~al.
\newblock Tuber: Tubelet transformer for video action detection.
\newblock In {\em Proceedings of the IEEE/CVF Conference on Computer Vision and Pattern Recognition}, pages 13598--13607, 2022.

\bibitem{wei2022CVPR}
Chen Wei, Haoqi Fan, Saining Xie, Chao-Yuan Wu, Alan Yuille, and Christoph Feichtenhofer.
\newblock Masked feature prediction for self-supervised visual pre-training.
\newblock In {\em Proceedings of the IEEE/CVF Conference on Computer Vision and Pattern Recognition}, pages 14668--14678, 2022.

\bibitem{faure2023WCACV}
Gueter~Josmy Faure, Min-Hung Chen, and Shang-Hong Lai.
\newblock Holistic interaction transformer network for action detection.
\newblock In {\em Proceedings of the IEEE/CVF Winter Conference on Applications of Computer Vision}, pages 3340--3350, 2023.

\bibitem{sun2018ECCV}
Chen Sun, Abhinav Shrivastava, Carl Vondrick, Kevin Murphy, Rahul Sukthankar, and Cordelia Schmid.
\newblock Actor-centric relation network.
\newblock In {\em Proceedings of the European Conference on Computer Vision (ECCV)}, pages 318--334, 2018.

\bibitem{stroud2020CVPR}
Jonathan Stroud, David Ross, Chen Sun, Jia Deng, and Rahul Sukthankar.
\newblock D3d: Distilled 3d networks for video action recognition.
\newblock In {\em Proceedings of the IEEE/CVF Winter Conference on Applications of Computer Vision}, pages 625--634, 2020.

\bibitem{shah2022CVPR}
Anshul Shah, Shlok Mishra, Ankan Bansal, Jun-Cheng Chen, Rama Chellappa, and Abhinav Shrivastava.
\newblock Pose and joint-aware action recognition.
\newblock In {\em Proceedings of the IEEE/CVF Winter Conference on Applications of Computer Vision}, pages 3850--3860, 2022.

\bibitem{song2019CVPR}
Lin Song, Shiwei Zhang, Gang Yu, and Hongbin Sun.
\newblock Tacnet: Transition-aware context network for spatio-temporal action detection.
\newblock In {\em Proceedings of the IEEE/CVF Conference on Computer Vision and Pattern Recognition}, pages 11987--11995, 2019.

\bibitem{pramono2019CVPR}
Rizard Renanda~Adhi Pramono, Yie-Tarng Chen, and Wen-Hsien Fang.
\newblock Hierarchical self-attention network for action localization in videos.
\newblock In {\em Proceedings of the IEEE/CVF International Conference on Computer Vision}, pages 61--70, 2019.

\bibitem{zhang2019CVPR}
Yubo Zhang, Pavel Tokmakov, Martial Hebert, and Cordelia Schmid.
\newblock A structured model for action detection.
\newblock In {\em Proceedings of the IEEE/CVF Conference on Computer Vision and Pattern Recognition}, pages 9975--9984, 2019.

\bibitem{li2020ECCV}
Yixuan Li, Zixu Wang, Limin Wang, and Gangshan Wu.
\newblock Actions as moving points.
\newblock In {\em Computer Vision--ECCV 2020: 16th European Conference, Glasgow, UK, August 23--28, 2020, Proceedings, Part XVI 16}, pages 68--84. Springer, 2020.

\bibitem{liu2021ACDnet}
Yu~Liu, Fan Yang, and Dominique Ginhac.
\newblock Acdnet: An action detection network for real-time edge computing based on flow-guided feature approximation and memory aggregation.
\newblock {\em Pattern Recognition Letters}, 145:118--126, 2021.

\bibitem{kumar2022CVPR}
Akash Kumar and Yogesh~Singh Rawat.
\newblock End-to-end semi-supervised learning for video action detection.
\newblock In {\em Proceedings of the IEEE/CVF Conference on Computer Vision and Pattern Recognition}, pages 14700--14710, 2022.

\bibitem{li2022SIVP}
Hui Li, Wenjun Hu, Ying Zang, and Shuguang Zhao.
\newblock Action recognition based on attention mechanism and depthwise separable residual module.
\newblock {\em Signal, Image and Video Processing}, pages 1--9, 2022.

\bibitem{nagrani2021NIPS}
Arsha Nagrani, Shan Yang, Anurag Arnab, Aren Jansen, Cordelia Schmid, and Chen Sun.
\newblock Attention bottlenecks for multimodal fusion.
\newblock {\em Advances in Neural Information Processing Systems}, 34:14200--14213, 2021.

\bibitem{arnab2021vivit_iccv}
Anurag Arnab, Mostafa Dehghani, Georg Heigold, Chen Sun, Mario Lu{\v{c}}i{\'c}, and Cordelia Schmid.
\newblock Vivit: A video vision transformer.
\newblock In {\em Proceedings of the IEEE/CVF international conference on computer vision}, pages 6836--6846, 2021.

\bibitem{patrick2021NIPS}
Mandela Patrick, Dylan Campbell, Yuki~M. Asano, Ishan Misra~Florian Metze, Christoph Feichtenhofer, Andrea Vedaldi, and João~F. Henriques.
\newblock Keeping your eye on the ball: Trajectory attention in video transformers.
\newblock In {\em Advances in Neural Information Processing Systems (NeurIPS)}, 2021.

\bibitem{kondratyuk2021movinets_cvpr}
Dan Kondratyuk, Liangzhe Yuan, Yandong Li, Li~Zhang, Mingxing Tan, Matthew Brown, and Boqing Gong.
\newblock Movinets: Mobile video networks for efficient video recognition.
\newblock In {\em Proceedings of the IEEE/CVF Conference on Computer Vision and Pattern Recognition}, pages 16020--16030, 2021.

\bibitem{yan2022CVPR}
Shen Yan, Xuehan Xiong, Anurag Arnab, Zhichao Lu, Mi~Zhang, Chen Sun, and Cordelia Schmid.
\newblock Multiview transformers for video recognition.
\newblock In {\em Proceedings of the IEEE/CVF Conference on Computer Vision and Pattern Recognition}, pages 3333--3343, 2022.

\end{thebibliography}

\begin{IEEEbiography}[{\includegraphics[width=1in,height=1.25in,clip,keepaspectratio]{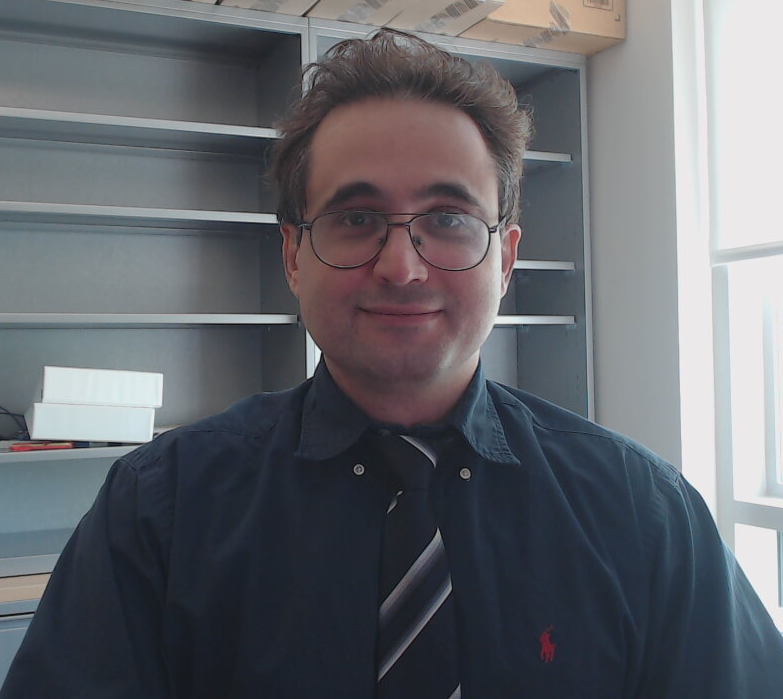}}]{Matthew Korban} (Senior Member, IEEE)
received his BSc and MSc degrees in Electrical Engineering from the University of Guilan, where he worked on sign language recognition in video. He received his Ph.D. in Computer Engineering from Louisiana State University. Matthew has led several groundbreaking projects, such as LSU CAVE2, one of the world's largest and most advanced Cave2 VR. He has published several publications in top-tier computer vision conferences and journals such as ECCV and Pattern Recognition. He is currently a Postdoc Research Associate at the University of Virginia, working with Prof. Scott T. Acton. His research interests include human action recognition, early action recognition, motion synthesis, and human geometric modeling.
\end{IEEEbiography}

\begin{IEEEbiography}[{\includegraphics[width=1in,height=1.25in,clip,keepaspectratio]{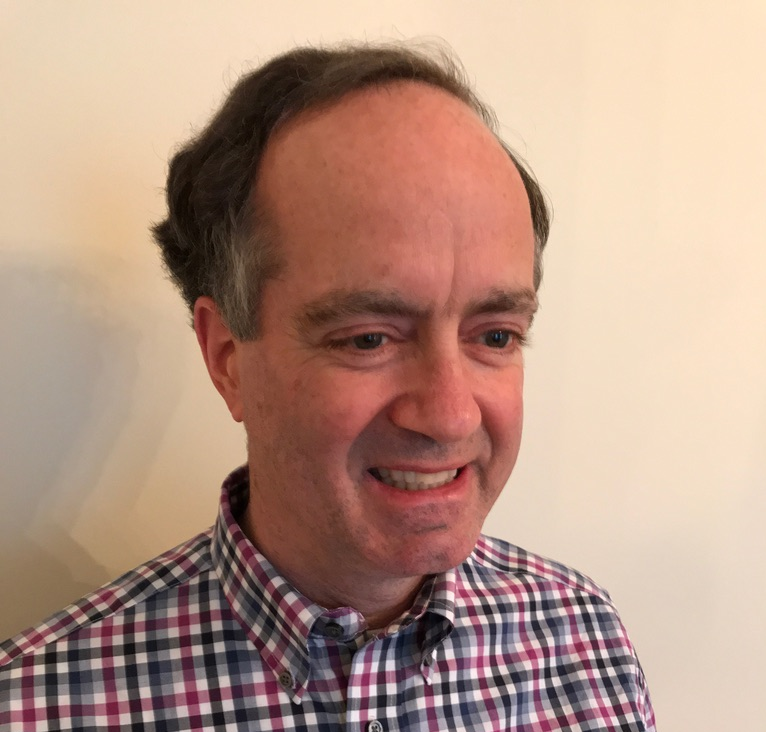}}]{Peter Youngs} is a Professor and Chair in the Department of Curriculum, Instruction, and Special Education at the University of Virginia. He studies how neural networks can automatically classify instructional activities in videos of elementary mathematics and reading instruction. He currently serves as co-editor of American Educational Research Journal.
\end{IEEEbiography}

\begin{IEEEbiography}[{\includegraphics[width=1in,height=1.25in,clip,keepaspectratio]{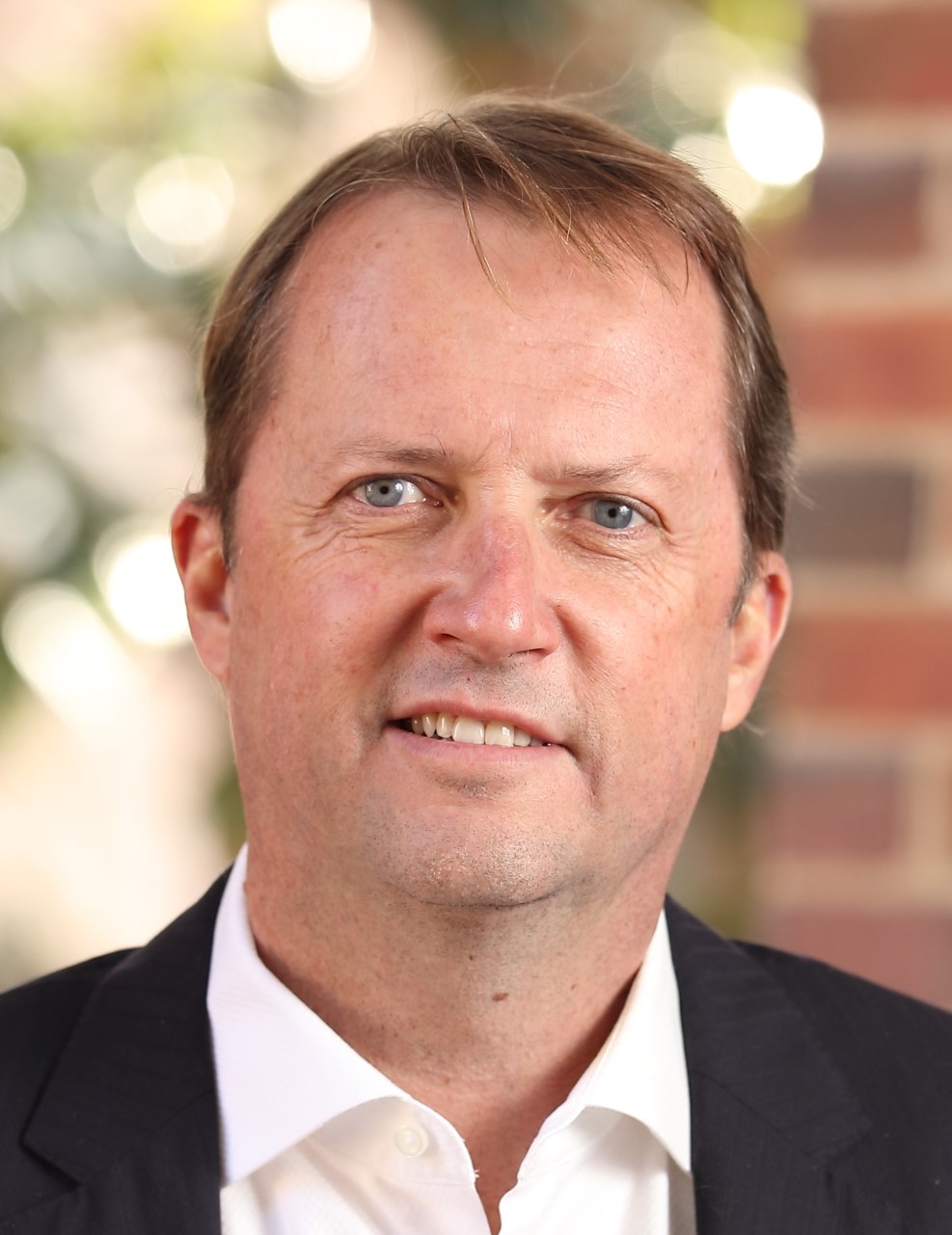}}]{Scott T. Acton}
is a Professor and Chair of Electrical and Computer Engineering at the University of Virginia. He is also appointed in Biomedical Engineering. For the last three years, he was Program Director in the Computer and Information Sciences and Engineering directorate of the National Science Foundation. He received the M.S. and Ph.D. degrees at the University of Texas at Austin, and he received his B.S. degree at Virginia Tech. Professor Acton is a Fellow of the IEEE “for contributions to biomedical image analysis.” Professor Acton’s laboratory at UVA is called VIVA - Virginia Image and Video Analysis. They specialize in biological/biomedical image analysis problems. The research emphases of VIVA include image analysis in neuroscience, tracking, segmentation, representation, retrieval, classification and enhancement. Recent theoretical interests include machine learning, active contours, partial differential equation methods, scale space methods, and graph signal processing. Professor Acton has over 300 publications in the image analysis area including the books Biomedical Image Analysis: Tracking and Biomedical Image Analysis: Segmentation. He was the 2018 Co-Chair of the IEEE International Symposium on Biomedical Imaging. Professor Acton was Editor-in-Chief of the IEEE Transactions on Image Processing (2014-2018)
\end{IEEEbiography}

\end{document}